\documentclass[10pt,twocolumn,letterpaper]{article}

\usepackage[pagenumbers]{cvpr} 

\usepackage[utf8]{inputenc}

\definecolor{cvprblue}{rgb}{0.21,0.49,0.74}
\usepackage[pagebackref,breaklinks,colorlinks,allcolors=cvprblue]{hyperref}
\usepackage[table]{xcolor}
\usepackage{booktabs}
\usepackage{pifont}
\usepackage{newtxtext,newtxmath}
\usepackage{xcolor}
\usepackage{algorithm}
\usepackage{algpseudocode}
\definecolor{algKeyword}{HTML}{162E93}   
\definecolor{algComment}{HTML}{0D9488}   
\definecolor{algFunc}{HTML}{B83280}      
\definecolor{algNet}{HTML}{553C9A}       
\definecolor{algOp}{HTML}{2B6CB0}        

\newcommand{\cgets}{\mathrel{\textcolor{algOp}{\gets}}}
\newcommand{\cle}{\mathrel{\textcolor{algOp}{\le}}}
\newcommand{\cge}{\mathrel{\textcolor{algOp}{\ge}}}

\def\paperID{11752} 
\def\confName{CVPR}
\def\confYear{2026}

\usepackage{graphicx}
\usepackage{capt-of}
\usepackage{booktabs}
\usepackage{multirow}
\usepackage{makecell}
\usepackage{graphicx} 
\usepackage[table]{xcolor}
\usepackage{colortbl}
\usepackage{subcaption} 
\usepackage{algorithm}
\usepackage{algpseudocode}

\graphicspath{{figs/}}
\newcommand{\TitleIcon}[1][3ex]{%
  \raisebox{-0.75ex}{\includegraphics[height=#1]{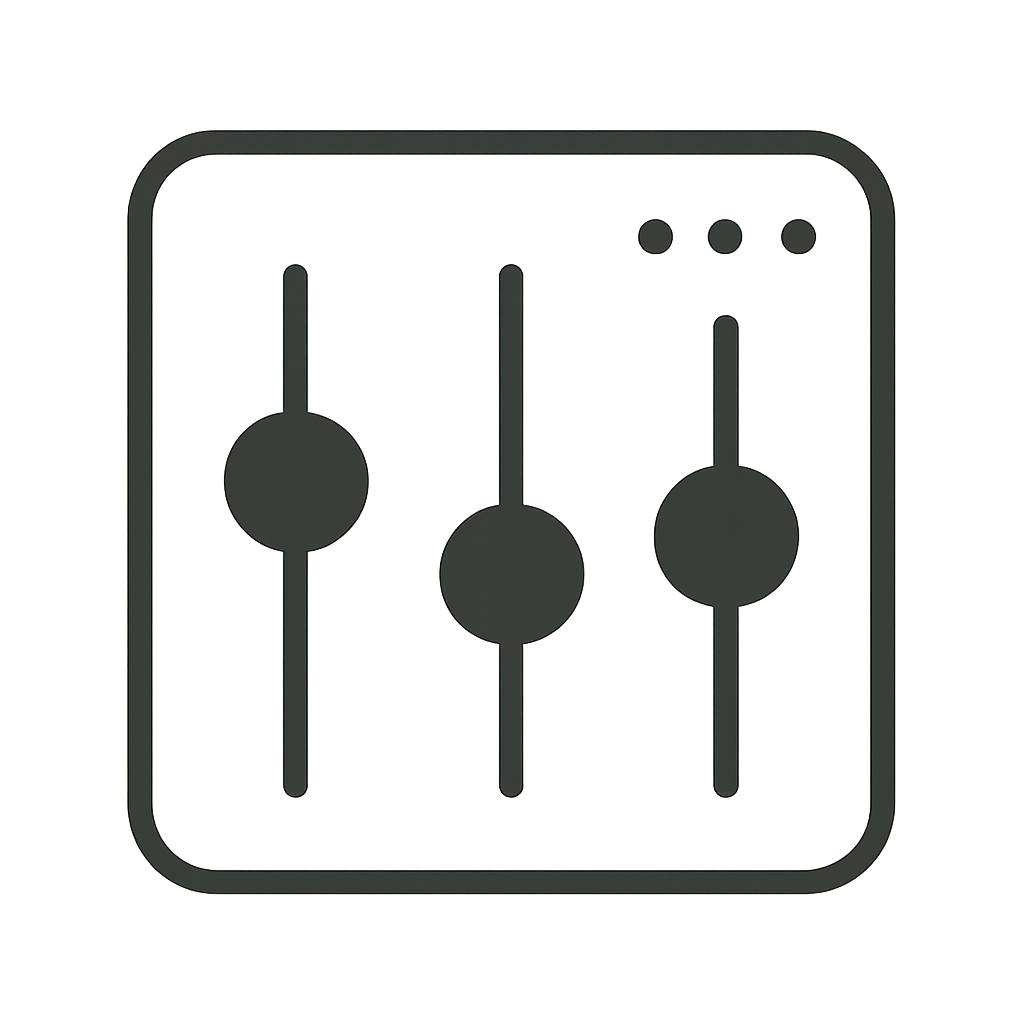}}%
}
\newcommand{\method}{\textbf{TunerDiT\;}}

\title{\texorpdfstring{\TitleIcon\ }{}TunerDiT: Training-free Progressive Steering of Diffusion Transformer for \\
Consistent Multi-Event Video Generation}

\author{
Ruotong Liao$^{1,3}$\thanks{Equal contribution.} \qquad 
Guowen Huang$^{2}$\footnotemark[1] \qquad 
Qing Cheng$^2$ \qquad Guangyao Zhai$^2$ \qquad Lei Zhang$^4$ \\ 
Xun Xiao$^5$ \qquad
Thomas Seidl$^{1,3}$ \qquad Daniel Cremers$^{2,3}$ \qquad Volker Tresp$^{1,3}$ \\[2.5ex]
$^1$ Ludwig Maximilian University of Munich \qquad $^2$ Technical University of Munich \\ 
$^3$ MCML \qquad $^4$ University of Hamburg \qquad $^5$ Huawei European Research Institute \\ \\
Project page: \url{https://tunerdit.github.io}
}

\makeatletter
\g@addto@macro\@maketitle{%
  \vspace{-0.8em}
  \begin{center}
    \setlength{\linewidth}{\textwidth}%
    \setlength{\hsize}{\textwidth}%
    \includegraphics[width=1.0\textwidth]{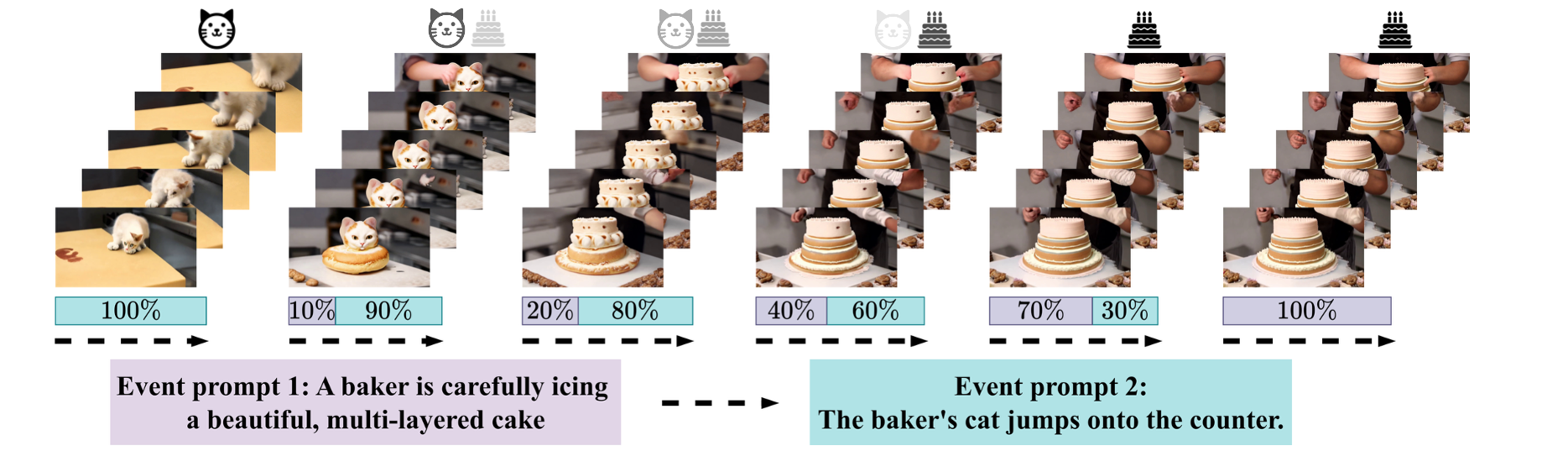}%
    \par\vspace{0.4em}%
    \captionof{figure}{\textbf{Different denoising steps utilize text inputs differently.} \textbf{TunerDiT} finds this insight by probing prompt conditioning of video diffusion models. When switching the input prompt at various fractions of the denoising steps (e.g., 0\%, 10\%, 20\%, 40\%, 70\%, 100\%), it is observed that early steps dominate the global layout while late steps refine fine-grained appearance and motion, revealing intrinsic turning points where text influence changes. Building on this insight, TunerDiT is a training-free, progressive steering method for video DiTs, yielding consistent multi-event videos with clear boundaries and smooth transitions.}%
    
    \label{fig:teaser}%
  \end{center}
  \vspace{0.6em}
}
\makeatother

\begin{document}
\maketitle
 


\begin{abstract}

Text-to-video (T2V) generation faces challenging questions when generating videos with long horizons containing multiple events
Inspired by the intrinsics of the diffusion process, we probe video diffusion transformers (DiTs) and uncover intrinsic turning points in the DiT denoising trajectory where conditioning text affects generation from global layout to fine-grained details. Building on this finding, we present \method, a simple yet effective progressive steering method that requires no additional training for multi-event generation. TunerDiT comprises two steering handles: (1) Event-Partitioned Masking that enforces event boundaries while allowing cross-event transition bands; (2) Cross-Event Prompt Fusion that injects neighboring event semantics for late-stage refinement. We contribute a self-curated prompt suite for benchmarking multi-event generation, i.e. \textit{\textsc{Meve}}. \method achieves state-of-the-art performance across 8 metrics and offers a tunable trade-off between video consistency and event separation, compared with other training-free methods. The improvement in text alignment increases with the event count, indicating a scaling possibility with increasing event count.


\end{abstract}


\section{Introduction}

\label{sec:intro}

Recent advances in text-to-video (T2V) generation have been largely driven by diffusion transformers (DiTs)~\cite{peebles2023scalablediffusionmodelstransformers, chen2024gentrondiffusiontransformersimage, xing2024surveyvideodiffusionmodels}
. Models such as CogVideoX~\citep{hong2022cogvideo, yang2024cogvideox} and OpenSora~\citep{lin2024open} adopt DiT-style architectures with spatial-temporal self-attention, 3D VAEs~\citep{nash2017shape}, and large-scale pretraining, achieving impressive visual quality on short clips for one good moment, which are insufficient for long-horizon, multi-event sceinarios required by many real-world applications~\citep{li2025worldmodelbench,kang2024far, cho2024sora, liu2025worldweaver, yang2025instadrive, liao2024videoinsta, liao2024gentkg}, such as long-horizon robotic planning, autonomous driving, interactive world models, etc. 
When naively repurposed for multi-event generation via direct concatenation of event prompts in time or text, existing methods exhibit clear, distinct failure patterns. The resulting synthesis either blends all events into a single muddled scene. The core desiderata of multi-event video generation are: (1) \textit{Event ordering}: clear event boundaries and correct temporal ordering; (2) \textit{Smooth transitions}: smooth handovers between events; (3) \textit{Semantic Consistency}: fine-grained semantic consistency across events bidirectionally between early and late events.

Only limited work brings attention to multi-event T2V generation that \cite{cai2025ditctrl, oh2024mevg, wang2023genlvideomultitextlongvideo, qi2025mask2ditdualmaskbaseddiffusion, qiu2024freenoisetuningfreelongervideo, liao2025eventsswitchmultieventvideo} directly address these questions. Achieving these requirements simultaneously has been proven computationally expensive~\cite{wu2025mind, kara2025shotadapter} and hard to scale to more events beyond the training scope. Other computationally light works achieve adequate but not satisfactory performance~\citep{wang2023gen, cai2025ditctrl, oh2024mevg}. However, looking inward at the denoising dynamics and cross-attention structure of DiTs indicates a more principled and computationally tractable approach that can simultaneously address the comprehensive core demands of multi-event generation.

Inspired by observations in text-to-image (T2I) diffusion that generation proceeds from coarse global structure to fine-grained details over the denoising trajectory~\citep{balaji2022eDiff-I, chen2024training, yang2023dynamic}, we hypothesize that Video DiTs exhibit analogous behavior in T2V generation, given the shared intrinsic properties of the diffusion process. Under this assumption, the temporal ordering of video events can be viewed as an additional dimension of fine-grained detail to be refined over the denoising process, rather than as a single monolithic structure.


To validate the hypothesis,  we first curate a benchmark suite \textit{\textsc{Meve}} that consists of multi-event prompts from multiple aspects. After systematically probing off-the-shelf DiT video generators by switching event conditioning at different denoising timestep fractions, we observe the resulting text-video similarity~\citep{kara2025shotadapter} as shown in Figure \ref{fig:teaser}. Consequently, we identify that text conditioning on early steps dominates the global video layout while text conditioning on late steps refines fine-grained appearance and motion, revealing the intrinsic turning point where text influence changes and is remarkably relatively stable within a model.




Building on this insight, we introduce \textbf{\method}, a progressive coarse-to-fine steering framework for multi-event T2V generation that operates \emph{without training}.
The idea is to first generate a shared layout of multiple events and refine inner event details at a certain later stage.
\method contains steerable control handles targeting different challenges mentioned above and exploits \emph{intrinsic turning points} in the DiT denoising process by intervening at the appropriate phase. Concretely, \method sets two steering handles to be activated according to a schedule:

\begin{itemize}
  \item \textbf{Cross-Event Prompt Fusion.} A gating scheme that conditions video latents on event prompts to enhance semantic awareness and coherence.
  \item \textbf{Event Partitioned Mask.} A diagonal mask that isolates events with connecting bands across events on DiT attention layers. The mask design restrains DiT's attention to enforce event boundaries and ensure smooth handovers.

\end{itemize}

Extensive experiments validate the effectiveness of \method. We achieve state-of-the-art results in multi-event video generation up to four events, with principal extensibility to more. \method consistently improves background consistency, identity consistency, text–video alignment, and transition smoothness over strong baselines. \method is performed at inference time and executed on a single A100 GPU, making our method cost-effective for real-world deployment. To summarize, our contributions are as follows:


\begin{itemize}
    \item We propose \textbf{\textsc{MEve}}, a benchmark prompt suite for multi-event video generation up to 4 events in multiple aspects. 
    \item We share the insight of \textbf{intrinsic turning point} from coarse to fine-grained generation in the Video DiT diffusion process.
    \item We propose \method, a tuner console-like steering approach for video diffusion transformers to generate multi-event consistent videos under a training-free setting.
\end{itemize}

\section{Related Work}

\paragraph{DiT-based Video Generation}
Recent works such as CogVideoX~\cite{hong2022cogvideo, yang2024cogvideox},
Mochi 1~\citep{genmo2024mochi}, Movie Gen~\citep{polyak2024movie}, and Hunyuan~\citep{sun2024hunyuan, li2024hunyuan}, VideoCrafter~\citep{chen2024videocrafter2}, etc, have adopted the DiT~\citep{peebles2023scalablediffusionmodelstransformers} architecture and achieved impressive results on generating one good moment. In this work, we build upon the open-source model OpenSora~\cite{lin2024open, li2024wf}. 
The architecture evolved from a Spatial-Temporal DiT (ST-DiT) structure to a dual-stream DiT. ST-DiT employs separate spatial and temporal attention blocks for frame-level and temporal processing, while the dual-stream DiT integrates image and text through parallel streams with unified attention mechanisms in one whole matrix.

\paragraph{Multi Event Video Generation}
Some works autoregressively generate videos based on the previous frames, which lose control of the global common sense of multi-events \cite{kim2024fifo, wang2023gen}.  Some make contributions to attentive mechanisms in DiT layers, e.g, introducing a temporal cross-attention layer that attends to different event scopes \cite{wu2025mind}, introducing transition tokens between cross-event text and video cross-attentions ~\cite{kara2025shotadapter},  or introducing symmetric masking on text-video cross-attention to ensure isolation among events at the cost of sharp scene transitions \cite{qi2025mask}, etc. These works demand significantly intensive computational resources and meticulously labeled video training data, yet they tend to deliver adequate but not exceptional results.

Consequently, some zero-shot methods are witnessed. They either focus on the guiding diffusion process based on last frame-aware latent initialization, while ignoring the impact regarding global consistency~\citep{oh2024mevg}, or on semantic priors on compositional layouts, but ignoring temporal orders~\cite{zhang2025magiccomp}, or on sharing subject-related attention by segmenting its binary mask~\citep{cai2025ditctrl}, which are better for background transition instead of subject-related motion change across events. However, the proposed \method is the first zero-shot method to tackle event ordering, smooth transition, and global semantic consistency together from a unified starting perspective of the intrinsics of the diffusion process and proposes a progressive steering method.
\section{The Intrinsic Turning Point in DiT}
\label{sec:intrinsic}

In this section, we first introduce the failure mode of general DiT-based Video Diffusion Models in generating multi-events, then introduce the hypothesis of the varying text conditioning effects at different diffusion step fractions in DiTs. To validate this hypothesis, we propose a new benchmark ~\textsc{Meve} and probe the intrinsic points in the DiT models, and show the resulting insight.
\subsection{Failure Modes in DiT-based models.}
Existing DiT video diffusion models face many issues like:

\textbf{(a) Event fusion.} All prompts are entangled into a single blended scene; Objects and actions from different events co-occur in most frames.\\[0.3em]
    \textbf{(b) Srambling order.} Events appear in random order or overlap with earlier ones, leading to ambiguous event ordering and even disappearing.\\[0.3em]
    \textbf{(c) Transition collapse.} Transitions are either abrupt frame-wise jumps or overly smooth morphs without clear event boundaries, harming event isolation.
  
\begin{figure*}[htbp!]
  \centering
  \begin{minipage}[b]{1.0\linewidth}
    \centering
    \includegraphics[width=\linewidth]{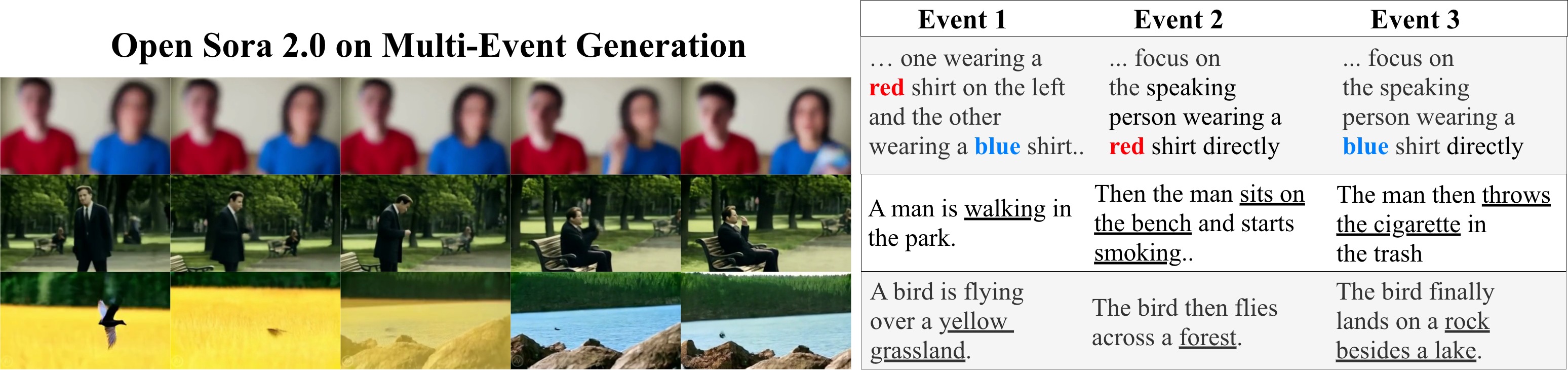}
  \end{minipage}\hfill
  \caption{DiT video diffusion models failed with multi-event prompting.}
  \label{fig:failure_modes}
\end{figure*}

\subsection{Benchmark, Hypothesis \& Validation: Intrinsic Turning Point.}

\begin{figure*}[t]
  \centering
  \small

  \begin{minipage}[t]{0.49\linewidth}
    \centering

    \begin{subfigure}[t]{\linewidth}
      \centering
      \footnotesize 
      \setlength{\tabcolsep}{5pt} 
      \begin{tabular}{lccc}
        \toprule
        \textbf{Category} & \textbf{View} & \textbf{Events/Prompt} & \textbf{\#Prompts} \\
        \midrule
        General            & 3rd        & 4 & 60 \\
        Motion Order       & 3rd        & 4 & 98 \\
        Human Identity     & 3rd        & 4 & 32 \\
        Complex Plot       & 3rd        & 4 & 60 \\
        Ego-Exo4D (paired) & 1st \& 3rd & 4 & 50 $\times$ 2 \\
        \bottomrule
      \end{tabular}
      \caption{Composition statistics per \textsc{MEve} category.}
      \label{tab:meve_overview}
    \end{subfigure}

    \vspace{1.0em}

    \begin{subfigure}[t]{\linewidth}
      \centering
      \scriptsize 
      \setlength{\tabcolsep}{3.5pt} 
      \begin{tabular}{lccccc}
        \toprule
        \multicolumn{1}{c}{Benchmark} &
        \multicolumn{1}{c}{\shortstack{Shot\\Counts}} &
        \multicolumn{1}{c}{\shortstack{Multi-\\Scene}} &
        \multicolumn{1}{c}{\shortstack{Multi-\\Event}} &
        \multicolumn{1}{c}{\shortstack{Real-world\\Reference}} &
        \multicolumn{1}{c}{\shortstack{View\\Control}} \\
        \midrule
        VBench        & 1 & $\times$     & $\times$     & $\times$     & $\times$     \\
        VBench++      & 1 & $\times$     & $\times$     & $\times$     & $\times$     \\
        VBench2.0     & 1 & $\times$     & $\times$     & $\times$     & $\times$     \\
        WorldScore    & 3 & $\checkmark$ & $\times$     & $\checkmark$ & $\checkmark$ \\
        MVBench       & 2 & $\checkmark$ & $\times$     & $\times$     & $\checkmark$ \\
        \textsc{MEve} & 4 & $\checkmark$ & $\checkmark$ & $\checkmark$ & $\checkmark$ \\
        \bottomrule
      \end{tabular}
      \caption{Comparison with existing video generation benchmarks.}
      \label{tab:benchmark_comparison}
    \end{subfigure}

  \end{minipage}
  \hfill
  \begin{minipage}[t]{0.49\linewidth}
    \centering

    \begin{subfigure}[t]{\linewidth}
      \centering
      \includegraphics[width=\linewidth]{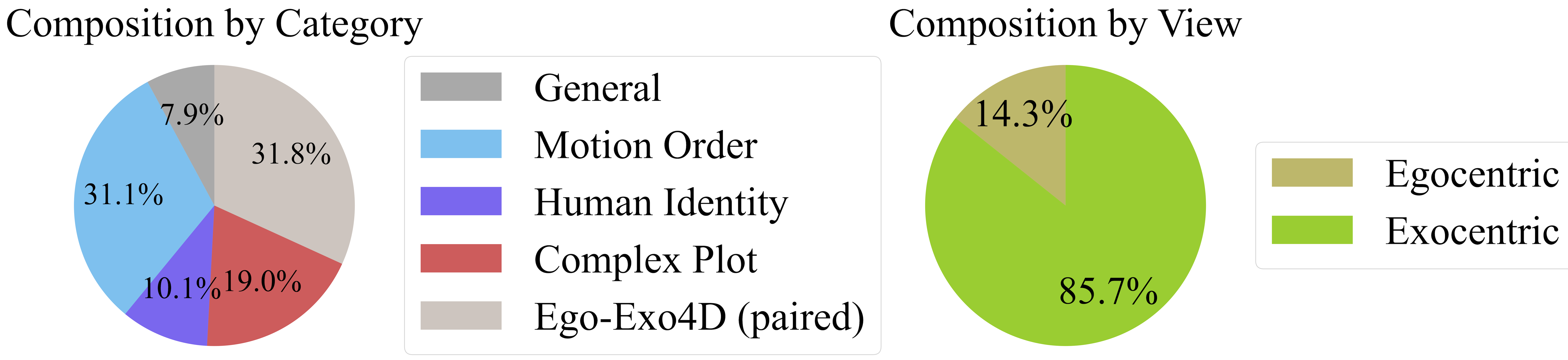}
      \caption{Category and view distribution of \textsc{MEve}.}
      \label{fig:meve_pies}
    \end{subfigure}

    \vspace{1.0em}

    \begin{subfigure}[t]{\linewidth}
      \centering
      \includegraphics[width=0.9\linewidth]{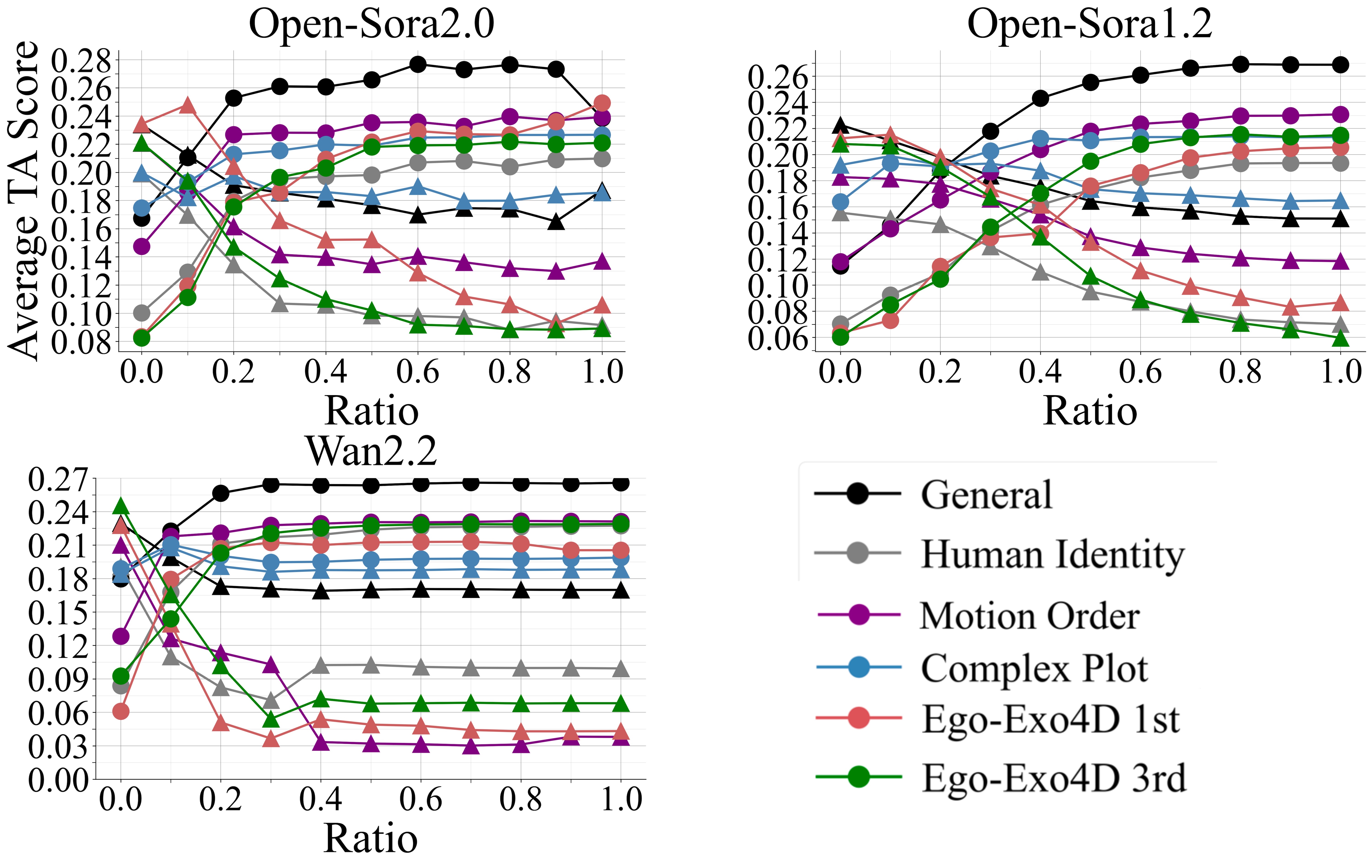}
      \caption{Experiment results on \textsc{MEve} for probing turning points.}
      \label{fig:tunerbutton}
    \end{subfigure}

  \end{minipage}

  \caption{\textsc{MEve} and turning-point analysis. (a,c) Dataset structure: category and view distribution statistics. (b) Position of \textsc{MEve} among existing video generation benchmarks. (d) Probing results revealing intrinsic turning points for event control in Video DiTs. }
  \label{fig:meve_fourpanel}
\end{figure*}

\paragraph{\textbf{Hypothesis: from T2I to T2V.}} Inspired by phenomena in text-to-image (T2I)~\citep{balaji2022eDiff-I, chen2024training, yang2023dynamic} that the diffusion process generates images from coarse to fine-grained, we hypothesize that the same happens for text-to-video (T2V) with Video DiTs due to the shared intrinsics of the diffusion process. If so, it is possible to treat the temporal dynamics of video events as an additional dimension of fine-grained details instead of a one-time generation. 

\paragraph{\textbf{Validation: Switching Denoising Fractions.}}
For simplicity in validating our hypothesis, we utilize two events for switching text conditioning during the iterative denoising process, noted by $\text{P}_1$ and $\text{P}_2$. As shown in Figure \ref{fig:teaser}, we change the input text to $\text{P}_2$ after a fixed percentage of denoising steps have been performed using $\text{P}_1$. Comparing results from left to right, we note that the text inputs have limited visible impact on the output when used in the first 10\% percent and gradually become more visible as the percentage goes up, especially in the appearance of the subject, which suggests text prompts affect only fine-grained details in the last denoising steps. As the percentage goes up continuously, the output shows influences fused from both prompts and gradually overrides the first text input's layout. From this observation, we find the denoiser utilizes text input differently at different noise levels, early steps dominate layouts, while the later steps refine details. Hence, we construct a benchmark to systematically probe the intrinsic turning point of DiTs.


\paragraph{\textbf{Benchmark: \textsc{Meve}}}
Existing T2V evaluation benchmarks focus on single-shot generation, dual-shot transitions, and seldom tackle multi-event prompts systematically~\cite{huang2024vbench, zheng2025vbench, huang2024vbenchcomprehensiveversatilebenchmark}. To systematically evaluate multi–event synthesis, we construct \textsc{MEve}, a prompt suite comprising multi-event prompts collected from three complementary sources, spanning up to four events per prompt. Sources include (1) LLM as a data synthesizer that generates per instructions~\cite{kara2025shotadapter}; (2) Prompts of diagnostic content that are expanded from VBench~2.0 categories~\cite{zheng2025vbench, zhang2024evaluationagent}; and (3) Prompts of viewpoint control that are paired in 1st view and 3rd views curated from video narrations in Ego–Exo4D~\citep{grauman2024egoexo4dunderstandingskilledhuman}. Refer to Appendix \ref{sec:app:dataset} for construction details. Data statistics are shown in Figure \ref{tab:meve_overview},\ref{tab:benchmark_comparison}.



\begin{figure*}[t]
  \centering
  \begin{subfigure}{0.22\linewidth}
    \includegraphics[width=0.9\textwidth]{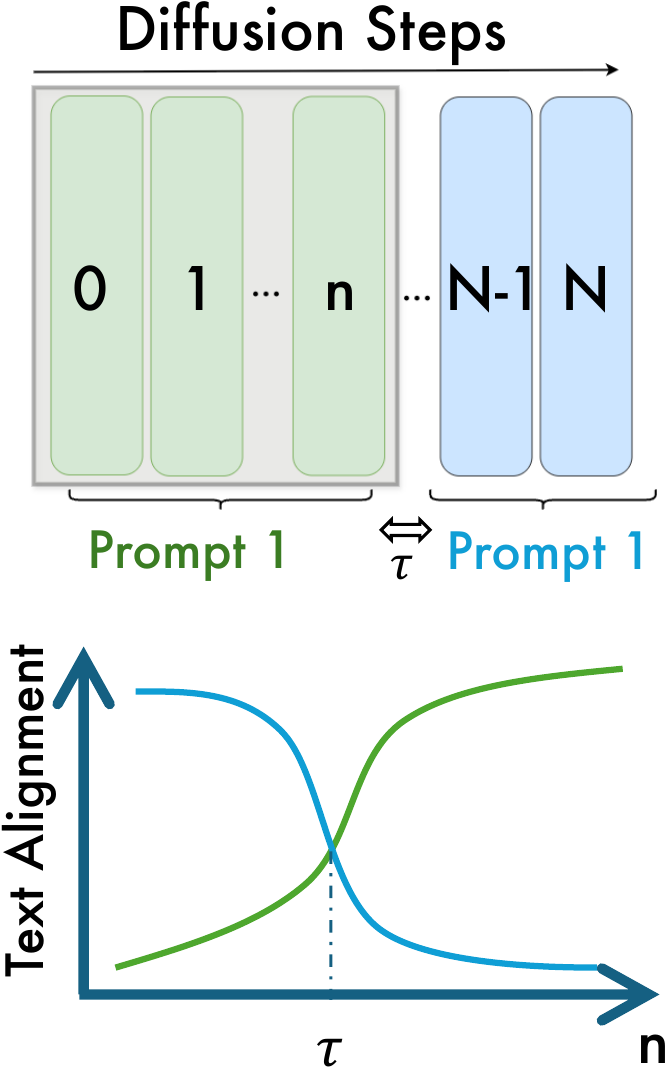}
    \caption{Probing the turning point.}
    \label{fig:main-short-a}
  \end{subfigure}
  \hfill
  \begin{subfigure}{0.77\linewidth}
    \includegraphics[width=1.0\textwidth]{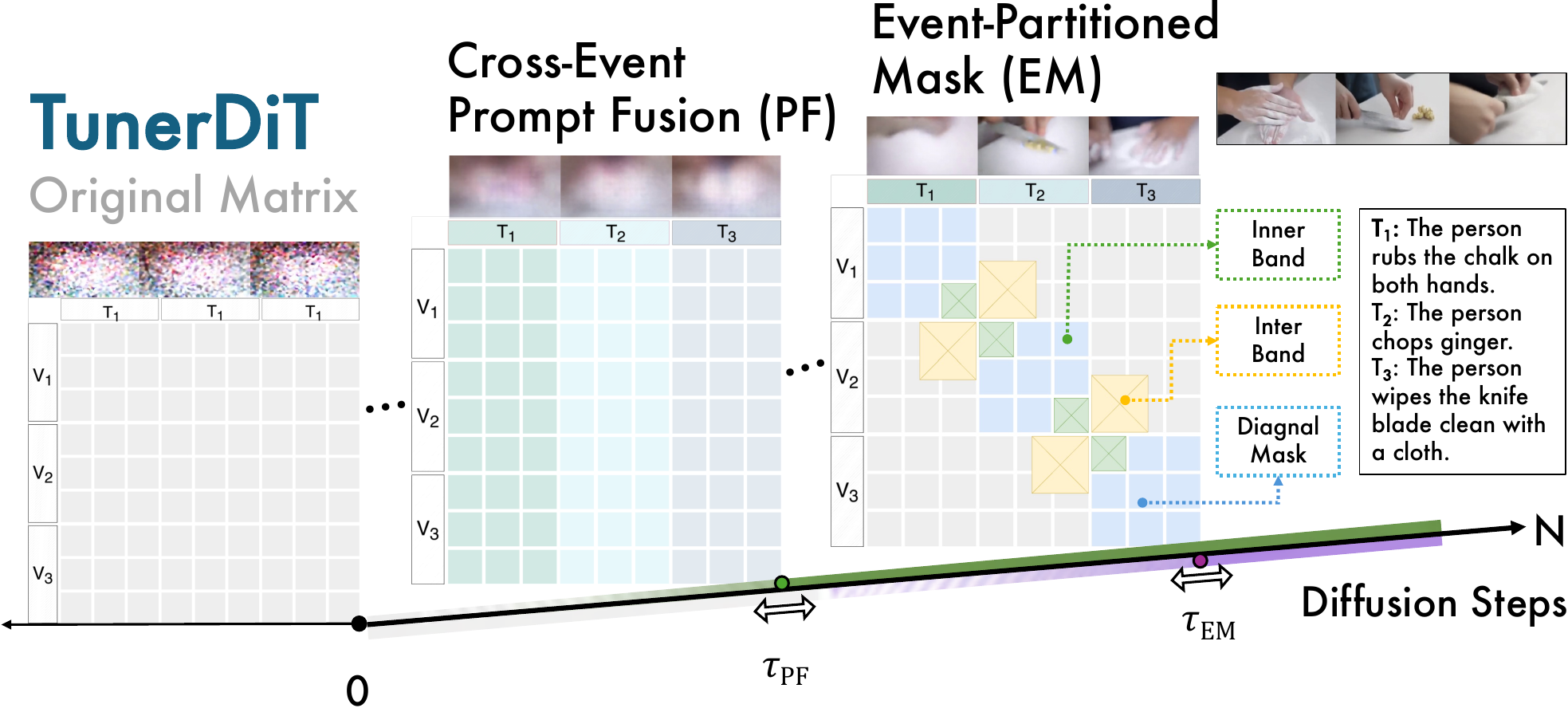}
    \caption{Turning point guided steering controls applied over denoising steps.}
    \label{fig:main-short-b}
  \end{subfigure}
\caption{
\textbf{\method} progressively steers multi-event generation over diffusion steps.
\emph{Cross-Event Prompt Fusion (PF)} first shares a common prompt to build a coherent layout, then gradually separates event prompts.
\emph{Event-Partitioned Mask (EM)} subsequently enforces event isolation via diagonal cross-attention blocks and introduces cross-event transition bands around boundaries, enabling smooth and semantically consistent handovers between events.
}

  \label{fig:short}
\end{figure*}

\paragraph{\textbf{Probing: from Coarse to Fine-grained.}}
As shown in Figure \ref{fig:main-short-a}, to probe exactly when the turning points from coarse to fine-grain controlling of text conditioning occur in denoising steps, we separate two event prompts $\text{P}_{1}$ and $\text{P}_{2}$ and parameterize the dual–prompt conditioning schedule in the following:

Let $x\in[0,1]$ be the fusion ratio and $n\in[0,N]$ the denoising step, the conditioning schedule be
\[
\mathrm{cond}(n)=
\begin{cases}
\text{P}_{1}, & n < xN,\\
\text{P}_{2}, & n \ge xN~,
\end{cases}
\]
so that the early diffusion steps, proportion $x$, are conditioned by $\text{P}_{1}$ and the late diffusion steps, proportion $1{-}x$, are conditioned by $\text{P}_{2}$.

The \textit{intrinsic turning point} $\tau$ is defined as the one whose step is the latest and closest to the observed intersection of text video alignment score in a group of paired video events, i.e. $\tau=\lfloor xN\rfloor$. The intuition behind this is that the step is where the DiTs hesitate to determine which is the dominant event layout. 
For steps previous to $\tau$ we condition on $\text{P}_{1}$; for steps starting from $\tau$ we condition on $\text{P}_{2}$. In practice, we sweep $x$ over a fixed grid
$\mathcal{X}=\{0,\,0.1,\,0.2,\,\ldots,\,1.0\} \quad (\text{i.e., } \Delta x=0.1).$ and we locate the $\tau$  when Text Alightment scores are intersecting for both prompts as the intrinsic turning points.

\paragraph{\textbf{The position of the intrinsic turning point.}} 


Surprisingly, this intrinsic turning point $\tau$ remains relatively stable for a given model, despite slight shifts across different event categories. Empirically, shown in Figure \ref{fig:teaser} and Figure \ref{fig:tunerbutton}, we observe that Text Alignment (TA) increases sharply within the early denoising steps. The turning point of the dominance of the second prompt comes in $x\in[0,0.3]$, followed by a broad plateau. This indicates that the temporal turning point is set early in diffusion denoising steps: exposing the model to a new event $\text{P}_2$ within the first $\sim 30\%$ steps is sufficient to trigger a shift, while later steps have a diminishing influence.
In other words, $\tau$ marks an internal phase boundary of the diffusion process where text conditioning transitions from structuring the coarse multi-event layout to refining high-frequency details. DiT-based video diffusion models implicitly separate layout planning and detail refinement along the denoising trajectory, and the location of this separation can be utilized as a controllable “tuning knob’’ for multi-event generation.

\section{Method: \method}

In this section, we present \method (Alg.~\ref{alg:tunerdit}), a training-free progressive steering framework that addresses event ordering, transition smoothness, and semantic-aware global consistency. \method follows the coarse-to-fine schedule revealed by the \textit{intrinsic turning-point} analysis (\S\ref{sec:intrinsic}) and operates in two phases: (1) \emph{Cross-Event Prompt Fusion} that first generates a shared global layout (§\ref{sec:PF}); and (2) an \emph{Event Partitioned Mask} that progressively separates events while connects events with smooth transitions (§\ref{sec:EM}).

\begin{algorithm}[t]
\caption{Progressive steering schedule of \method}
\label{alg:tunerdit}
\begin{algorithmic}[1]
\Require Event embeddings $\{T_1,\dots,T_E\}$, video embedding $V$, \\ total denoising steps $N$
\Require Candidate ratios $x_{\mathrm{EM}}^{\text{list}}$, $x_{\mathrm{PF}}^{\text{list}}$
\Ensure Progressive DiT attention with prompt fusion and event masking

\State \Comment{\textcolor{algComment}{\textbf{EM}: select the best event-partitioning ratio}}
\State $x_{\mathrm{EM}} \cgets \arg\max_{x \in x_{\mathrm{EM}}^{\text{list}}} \textcolor{algFunc}{\mathrm{TA}_{\mathrm{EM}}}(x)$
\State $\tau_{\mathrm{EM}} \cgets \lfloor N \cdot x_{\mathrm{EM}} \rfloor$

\State \Comment{\textcolor{algComment}{\textit{\textbf{PF}: select the best prompt-fusion ratio}}}
\State $x_{\mathrm{PF}} \cgets \arg\max_{x \in x_{\mathrm{PF}}^{\text{list}}} \textcolor{algFunc}{\mathrm{TA}_{\mathrm{PF}\mid \mathrm{EM}}}(x)$
\State $\tau_{\mathrm{PF}} \cgets \lfloor N \cdot x_{\mathrm{PF}} \rfloor$

\For{$n = 0$ \textcolor{algKeyword}{\textbf{to}} $N-1$}
    \State \Comment{\textcolor{algComment}{\textbf{PF}: activation of cross-event prompt fusion}}
    \If{$n \cle \tau_{\mathrm{PF}}$}
        \State $T(n) \cgets \textcolor{algFunc}{\mathrm{concat}}(T_1, T_1, \dots, T_1)$
    \Else
        \State $T(n) \cgets \textcolor{algFunc}{\mathrm{concat}}(T_1, T_2, \dots, T_E)$
    \EndIf

    \State \Comment{\textcolor{algComment}{\textbf{EM}: activation of event-partitioned mask}}
    \If{$n \cge \tau_{\mathrm{EM}}$}
        \State $D(n) \cgets \textcolor{algFunc}{\mathrm{GetEventMask}}(E, V)$
    \Else
        \State $D(n) \cgets \textcolor{algOp}{\varnothing}$
    \EndIf

    \State \Comment{\textcolor{algComment}{\textbf{DiT update}: perform steered conditioning}}
    \State $V \cgets \textcolor{algNet}{\mathrm{DiTAttention}}(V, T(n), D(n))$
\EndFor
\end{algorithmic}
\end{algorithm}

\subsection{Cross-Event Prompt Fusion}
\label{sec:PF}

Having the intrinsic turning point $\tau^\star$ probed for a model, \method first sets up a $\tau_{PF}$ as a tunable gating hyperparameter near to $\tau^\star$ for prompt fusion. Intuitively, the aim is to exploit the \emph{early} denoising steps before $\tau_{PF}$ to establish a coherent global layout before forcing the DiTs to distinguish events after $\tau_{PF}$. Let $E$ be the number of events and $P_e$ the prompt of event $e$ with text embeddings:
\[
T_e = f_{\text{text}}(P_e) \in \mathbb{R}^{m_e \times d}, \quad e=1,\dots,E,
\]
and concatenate them into $T=[T_1;\dots;T_E]$.

\textbf{Temporal gating of PF.} For diffusion step $n \in \{1,\dots,N\}$ \method constructs the effective text tokens $\widehat T_e(n)$ as
\begin{equation}
\widehat T_e(n) =
\begin{cases}
T_1, & n < \tau_{\text{PF}}, \\
T_e, & n \ge \tau_{\text{PF}},
\end{cases}
\label{eq:pf-schedule}
\end{equation}
and $\widehat T(n) = [\widehat T_1(n);\dots;\widehat T_E(n)]$.

Thus, in the \emph{coarse layout diffusion phase} when $n < \tau_{\text{PF}}$, all video tokens are conditioned on the first-event prompt $P_1$, encouraging a globally consistent scene and camera layout shared across the whole sequence. After the gate $\tau_{\text{PF}}$, the model starts to differentiate between events content details by conditioning video embeddings on their own event text embeddings, effectively “unfusing’’ the prompts and handing control over to event-specific conditioning.



\subsection{Event-Partitioned Diagonal Mask}
\label{sec:EM}

After PF builds a shared layout, \method activates the \emph{Event-Partitioned Diagonal Mask} (EM) to separate events while preserving smooth transitions. 
EM is applied after the denoising step $\tau_{\mathrm{EM}}$ shown in Algorithm~\ref{alg:tunerdit}. 
Let the video tokens be divided into $E$ consecutive event segments
\[
V=[V_1;\dots;V_E],
\]
and let the text tokens be divided into the corresponding event prompts
\[
T(n)=[T_1(n);\dots;T_E(n)].
\]
Here, $V_e$ and $T_e(n)$ denote the video and text tokens associated with event $e$.

\paragraph{Event-partitioned diagonal mask.}
The diagonal mask keeps each event segment focused on its own event prompt. 
Specifically, for video tokens in segment $V_e$, EM mainly allows attention to the matching text tokens $T_e(n)$ and suppresses attention to unrelated event prompts. 
This produces an event-wise diagonal attention pattern:
\[
D_{\mathrm{diag}}
=
\operatorname{diag}
\big(
V_1 \leftrightarrow T_1,\,
V_2 \leftrightarrow T_2,\,
\dots,\,
V_E \leftrightarrow T_E
\big),
\]
where $D_{\mathrm{diag}}$ denotes the binary-pass attention map for event-matched video--text pairs. 
This prevents semantic leakage across events and helps preserve the event order.

\paragraph{Transition bands.}
Transition bands allow neighboring events to communicate only near their boundary. 
A purely diagonal mask can create hard cuts between adjacent events, so for each boundary $e \rightarrow e+1$, we augment the diagonal structure with narrow rectangular bands. 
We use two types of bands $D_{\mathrm{band}}$: an \emph{inner-event} band $D_{\mathrm{inner}}^{v \leftarrow v}$ in video self-attention, and \emph{inter-event} bands $D_{\mathrm{inter}}^{v \leftarrow t}$ and $D_{\mathrm{inter}}^{t \leftarrow v}$ in video--text cross-attention. 
The inter-event bands allow boundary video tokens to attend to neighboring event text tokens, while the inner-event band lets video latents on both sides of the boundary exchange information in the latent space.

The band width is controlled by a ratio $r \in (0,1)$. 
For event $e$ with $|V_e|$ video tokens and $|T_e|$ text tokens, we set
\[
h_e^{vv} = h_e^{vt} = \lfloor r |V_e| \rfloor,
\qquad
w_e^{vt} = \lfloor r |T_e| \rfloor .
\]
Here, $h_e^{vv}$ and $h_e^{vt}$ determine the number of video tokens near the boundary that participate in the transition, and $w_e^{vt}$ determines the number of neighboring text tokens visible in cross-attention. 
Larger $r$ allows broader boundary neighborhoods to exchange information, while smaller $r$ restricts the interaction to a tighter transition region.


\paragraph{Temporal gating of EM.}
EM is disabled before $\tau_{\mathrm{EM}}$ and enabled only after the global layout has stabilized. 
Before $\tau_{\mathrm{EM}}$, all video tokens can attend to all text tokens. 
After $\tau_{\mathrm{EM}}$, only the event-wise diagonal blocks and transition bands are allowed:
\[
D^{v \leftarrow t}_{\mathrm{EM}}(n)
=
\begin{cases}
\mathbf{1}, & n < \tau_{\mathrm{EM}},\\
D_{\mathrm{diag}} \vee D_{\mathrm{band}}, & n \ge \tau_{\mathrm{EM}},
\end{cases}
\]
where $\vee$ denotes element-wise logical OR. 
The additive attention mask is then
\[
M^{v \leftarrow t}_{\mathrm{EM}}(n)
=
(-\infty)\cdot
\left(
\mathbf{1}
-
D^{v \leftarrow t}_{\mathrm{EM}}(n)
\right),
\]
which is added to the attention logits before the softmax. 
For text-to-video attention, we use the transpose of the same mask. 
For video self-attention, we apply the same event-wise blocking rule and keep boundary bands between adjacent video segments. 
Thus, EM isolates unrelated events while preserving enough boundary context for transitions.

\label{sec:metho}

\section{Experiments}

\begin{table*}[t]
\centering
\small
\caption{Quantitative comparison and preference-aligned evaluation. (a) Quantitative metrics across \{TA, TIS, BC, IC, CSCV\} and varying shot numbers \{2, 3, 4\}. (b) VLM-as-a-judge EI/TVA and human user study scores.}
\label{tab:main_and_vla}

\begin{subtable}{\textwidth}
\centering
\caption{Quantitative comparison of different models across five metrics. Metrics with the highest value are highlighted in \textbf{bold} and the second best are \underline{underlined}.}
\label{tab:main_results_resized}
\resizebox{\textwidth}{!}{%
\begin{tabular}{l|ccc|ccc|ccc|ccc|ccc}
\hline
\multicolumn{1}{c|}{} &
\multicolumn{3}{c|}{\textbf{TA$\uparrow$}} &
\multicolumn{3}{c|}{\textbf{TIS$\uparrow$}} &
\multicolumn{3}{c|}{\textbf{BC$\uparrow$}} &
\multicolumn{3}{c|}{\textbf{IC$\uparrow$}} &
\multicolumn{3}{c}{\textbf{CSCV$\uparrow$}} \\
\hline
\textbf{Shot Number}  & \textbf{2} & \textbf{3} & \textbf{4} &
                       \textbf{2} & \textbf{3} & \textbf{4} &
                       \textbf{2} & \textbf{3} & \textbf{4} &
                       \textbf{2} & \textbf{3} & \textbf{4} &
                       \textbf{2} & \textbf{3} & \textbf{4} \\
\hline
\multicolumn{16}{l}{\textbf{Zero-shot Methods}} \\
\hline
MEVG      & 0.201 & 0.206 & 0.205 & 0.271 & 0.270 & 0.272 & 0.228 & 0.249 & 0.270 & 0.269 & 0.270 & 0.270 & 0.688 & 0.703 & 0.707 \\
DiTCtrl   & 0.186 & 0.207 & \underline{0.216} & 0.259 & 0.271 & 0.278 & 0.303 & 0.377 & 0.394 & 0.280 & 0.354 & 0.389 & 0.826 & 0.819 & 0.803 \\
FreeNoise & 0.197 & 0.199 & 0.206 & 0.272 & 0.267 & 0.275 & 0.275 & 0.401 & 0.431 & 0.273 & 0.372 & 0.428 & 0.732 & 0.743 & 0.748 \\
\hline
\multicolumn{16}{l}{\textbf{Ours}} \\
\hline
\method{} Wan2.2        & 0.201 & 0.211 & 0.211 & 0.273 & 0.277 & 0.279 & \textbf{0.619} & \textbf{0.575} & \textbf{0.669} & \textbf{0.516} & \textbf{0.512} & \textbf{0.660} & 0.831 & \underline{0.840} & 0.830 \\
\method{} Open\mbox{-}Sora 1.2 & \underline{0.202} & \underline{0.211} & 0.213 & \underline{0.277} & \textbf{0.284} & \underline{0.281} & \underline{0.508} & 0.472 & \underline{0.496} & \underline{0.452} & 0.452 & 0.460 & \underline{0.848} & 0.839 & \underline{0.844} \\
\method{} Open\mbox{-}Sora 2.0 & \textbf{0.210} & \textbf{0.213} & \textbf{0.219} & \textbf{0.280} & \underline{0.277} & \textbf{0.287} & 0.501 & \underline{0.532} & 0.496 & 0.411 & \underline{0.488} & \underline{0.466} & \textbf{0.866} & \textbf{0.883} & \textbf{0.854} \\
\hline
\end{tabular}
}
\end{subtable}

\vspace{0.75em}

\begin{subtable}{\textwidth}
\centering
\caption{Evaluation metrics with human preference alignment. Left: VLM-as-a-judge for Event Isolation and Text-Video Alignment; right: human user study scores (18 people).}
\label{tab:merged_vla_human}

\begin{minipage}[t]{0.44\textwidth}
    \centering
    \footnotesize 
    \setlength{\tabcolsep}{6pt} 
    \begin{tabular}{lcc}
        \toprule
        \textbf{Name} & \textbf{EI} & \textbf{TVA} \\
        \midrule
        \multicolumn{3}{l}{\textbf{Zero-shot Methods}} \\
        MEVG           & 0.435 & 1.375 \\
        FreeNoise      & 0.436 & 1.400 \\
        DitCtrl        & 0.375 & 1.425 \\
        \midrule
        \multicolumn{3}{l}{\textbf{Ours}} \\
        \method{} Wan 2.2        & 0.474 & 1.503 \\
        \method{} Open-Sora 1.2  & 0.499 & 1.492 \\
        \method{} Open-Sora 2.0  & \textbf{0.572} & \textbf{1.533} \\
        \bottomrule
    \end{tabular}
\end{minipage}
\hfill
\begin{minipage}[t]{0.52\textwidth}
    \centering
    \footnotesize 
    \setlength{\tabcolsep}{6pt}
    \begin{tabular}{lcccc}
        \toprule
        \textbf{Model} & \textbf{Q1} & \textbf{Q2} & \textbf{Q3} & \textbf{Q4} \\
        \midrule
        \multicolumn{5}{l}{\textbf{Zero-shot Methods}} \\
        MEVG      & 1.82 & 1.91 & 1.82 & 2.05 \\
        FreeNoise & 1.99 & 2.01 & 1.99 & 1.79 \\
        DiTCtrl   & 2.11 & 2.03 & 2.18 & 2.35 \\
        \midrule
        \multicolumn{5}{l}{\textbf{Ours}} \\
        \method{} Wan 2.2  & 3.01 & 2.97 & 2.82 & 3.30 \\
        \method{} OpenSora 1.2  & 2.66 & 2.63 & 2.62 & 2.94 \\
        \method{} Open-Sora 2.0 & \textbf{3.16} & \textbf{3.03} & \textbf{3.03} & \textbf{3.34} \\
        \bottomrule
    \end{tabular}
\end{minipage}

\end{subtable}

\end{table*}
\definecolor{lightgray}{gray}{0.45}
\definecolor{lightgreen}{HTML}{E2F0D9}
\definecolor{green}{HTML}{007E6E}

\subsection{Emperimental Setup}
\paragraph{\textbf{Implementation Details}}
In this work, we apply \method{} on the representative large open-source models OpenSora 1.2~\cite{lin2024open}, OpenSora 2.0 ~\cite{li2024wf}, and Wan 2.2~\cite{wan2025}, spanning across different DiT structures, e.g., unified attention and factorized attention. Multi-event generation is evaluated on the proposed \textsc{MEve} dataset in a training-free setting.
To ensure a fair evaluation, each model generates videos using its native supported frame count and resolution specifications when provided with identical text prompts. For more information regarding mask sizes, turning point indices, etc., please refer to Appendix~\ref{sec:appendix_implementation}.

\paragraph{\textbf{Baselines}}
\label{sec:baseline}
To comprehensively evaluate the proposed framework, we benchmarked it against a diverse set of recent text-to-video methods covering both single-event and multi-event generation paradigms.
For open-source T2V large models, we included Open-Sora 1.2/2.0 as strong diffusion-transformer backbones representing large-scale open models.
We further compared with zero-shot baselines for fair comparison, i.e., MEVG~\citep{oh2024mevg}, 
DiTCtrl~\citep{cai2025ditctrl}, 
and FreeNoise~\citep{qiu2024freenoisetuningfreelongervideo}.

\paragraph{\textbf{Metrics}}
We evaluate multi-event generation from three complementary perspectives:
(1) automatic embedding/segmentation-based metrics adapted from single-shot T2V protocols;
(2) VLM-as-judge scores; and (3) human evaluations.

\textbf{(1) Embedding-based automatic metrics.}
Following commonly used metrics in single–shot T2V evaluation~\cite{wu2025mind}, their multi-shot adaptations in~\cite{kara2025shotadapter}, and additionally proposed by~\cite{cai2025ditctrl}, we report five automatic metrics:
\begin{itemize}[leftmargin=*]
  \item \textbf{Text Alignment (TA).}
  We assess text–video alignment by computing the cosine similarity between event-level text features and video features extracted by ViCLIP~\cite{wang2024internvidlargescalevideotextdataset}, and then averaging across events.

  \item \textbf{Identity Consistency (IC).}
  We quantify subject consistency by segmenting foreground entities with YOLOv11~\cite{yolo11_ultralytics} at the middle frame of each event and computing the average DINOv2~\cite{oquab2024dinov2learningrobustvisual} embedding similarity across events.

  \item \textbf{Background Consistency (BC).}
  We measure scene continuity by segmenting background regions with SAM2~\cite{ravi2024sam2segmentimages} and computing DINOv2 embedding similarity between midpoint frames of different events.

  \item \textbf{Text–Image Similarity (TIS).}
  We report CLIP similarity~\cite{radford2021learningtransferablevisualmodels} between the prompt and uniformly sampled frames from the generated video, following the TIS protocol of VBench 2.0~\cite{huang2023vbenchcomprehensivebenchmarksuite}. This reflects global semantic alignment of the full multi-event clip.

  \item \textbf{Clip Similarity Coefficient of Variation (CSCV).}
  CSCV computes the coefficient of variation of CLIP similarities across temporally sampled frames: it captures how smoothly the alignment score evolves across events, with higher CSCV indicating more stable, temporally coherent prompt adherence: 
  $ s_i = \mathbf{x}_i^\top \mathbf{x}_{i+1}, \quad i = 1,\ldots,n-1 $, with $\text{\texttt{CSCV}} = \frac{1}{1 + \lambda \cdot \frac{\sigma(s)}{\mu(s)}}\,$.
\end{itemize}

\textbf{(2) VLM-as-judge metrics.}
Beyond hand-crafted metrics, we further propose to evaluate videos generation with a vision–language model (VLM)-as-a-judge. For each generated video, we uniformly sample frames and query Gemini-2.5-Flash\cite{comanici2025gemini25pushingfrontier} to produce a caption for each frame and an overall summary of the clip. Based on captions, we define:
\begin{itemize}[leftmargin=*]
  \item \textbf{Event Isolation (EI).}
  From the summary caption, Gemini-2.5-Flash predicts the number of distinct events in the video. We compare this prediction with the ground-truth event count using a binary indicator of correctness and report the average accuracy over all videos.

  \item \textbf{Text–Video Alignment (TVA).}
  We encode both the ground-truth multi-event prompt and the VLM-generated summary caption with Sentence-BERT\cite{reimers2019sentence}, compute their cosine similarity, and average over videos. TVA thus reflects how well the overall video semantics match the intended multi-event description from a language perspective.
\end{itemize}

\textbf{(3) Human evaluation.}
Finally, we conduct a user study of 18 persons on a 5-point Likert scale, where annotators compare different models on four aspects: (Q1) overall video preference, (Q2) motion naturalness, (Q3) transition smoothness across events, and (Q4) text–video alignment. Detailed protocol, interface, annotator instructions, and additional statistics are provided in Appendix~\ref{sec:app:humaneval}.

\subsection{Experiment Results: Quantitative Analysis}
%



\begin{figure*}
  \centering
  \includegraphics[width=1.0\textwidth]{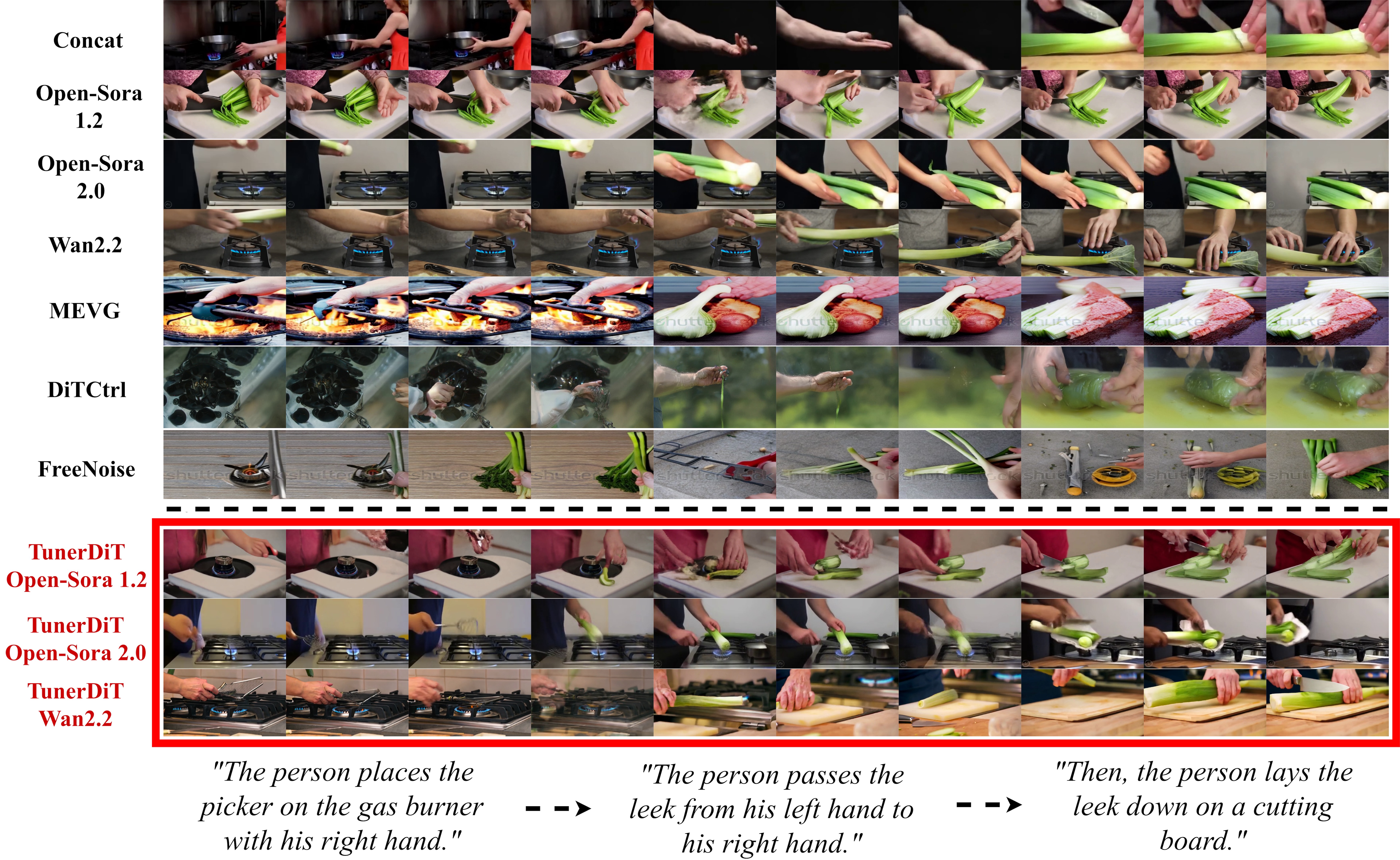}
  \caption{Qualitative comparison to various open-source base models (Open-Sora 1.2, Open-Sora 2.0 and Wan 2.2), direct concatenation of individual events (Concat), zero-shot multi-event baselines (MEVG, DitCtrl, FreeNoise).  Our methods achieve the best performance in event ordering, transition smoothness, and semantic aware consistency.}
  \label{fig:quali}
\end{figure*}



\paragraph{How is the TunerDiT steering quality of multi-event generation?}
We summarize four findings from Table~\ref{tab:main_results_resized}:

\textbf{\emph{State of the art over existing baselines.}} Our method attains the highest overall performance across multi-event settings compared with zero-shot baselines on all eight metrics spanning from two to four event settings. Considering open-source base models, we are able to surpass text-video alignment at longer horizons (TA, TIS), remain comparable results on smoothness (CSCV), and achieve the highest consistency value (BC, IC).

\textbf{\emph{Best human-aligned preference performance.}} 
As shown in Table~\ref{tab:merged_vla_human}, \method{} (Open-Sora~2.0) achieves the best VLM-as-a-judge scores, with the highest VLM-as-a-judge performance
(EI $=0.572$, TA $=1.533$), and simultaneously attains the top human preference scores across all four questions (Q1–Q4). 
Compared to the strongest zero-shot baseline DiTCtrl, this corresponds to absolute gains of $+1.05$, $+1.00$, $+0.85$, and $+0.99$ on overall quality, motion naturalness, transition smoothness, and text–video alignment, respectively. These results indicate that \method{} not only improves multi-event generation in automatic evaluation metrics but also aligns best with human judgments of video quality and coherence.


\subsection{Ablation Study}
\label{subsec:ablation}



\begin{figure}[htbp!]
    \centering
    \includegraphics[width=\linewidth]{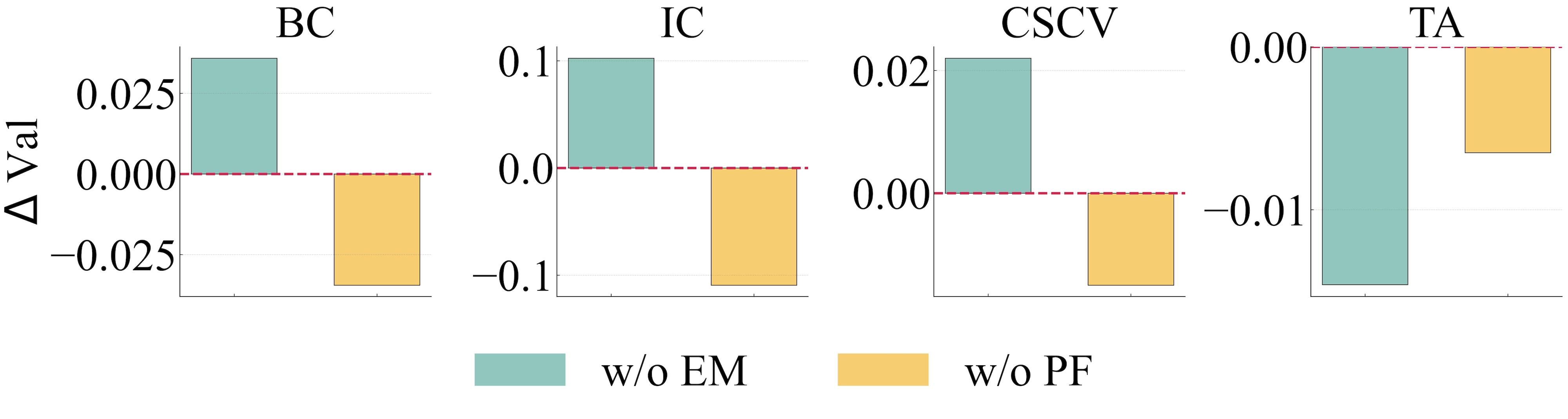}
    \caption{Per-component ablation results of \method{}.}
    \label{fig:ablation_components}
\end{figure}

\begin{figure}[htbp!]
    \centering
    \includegraphics[width=\linewidth]{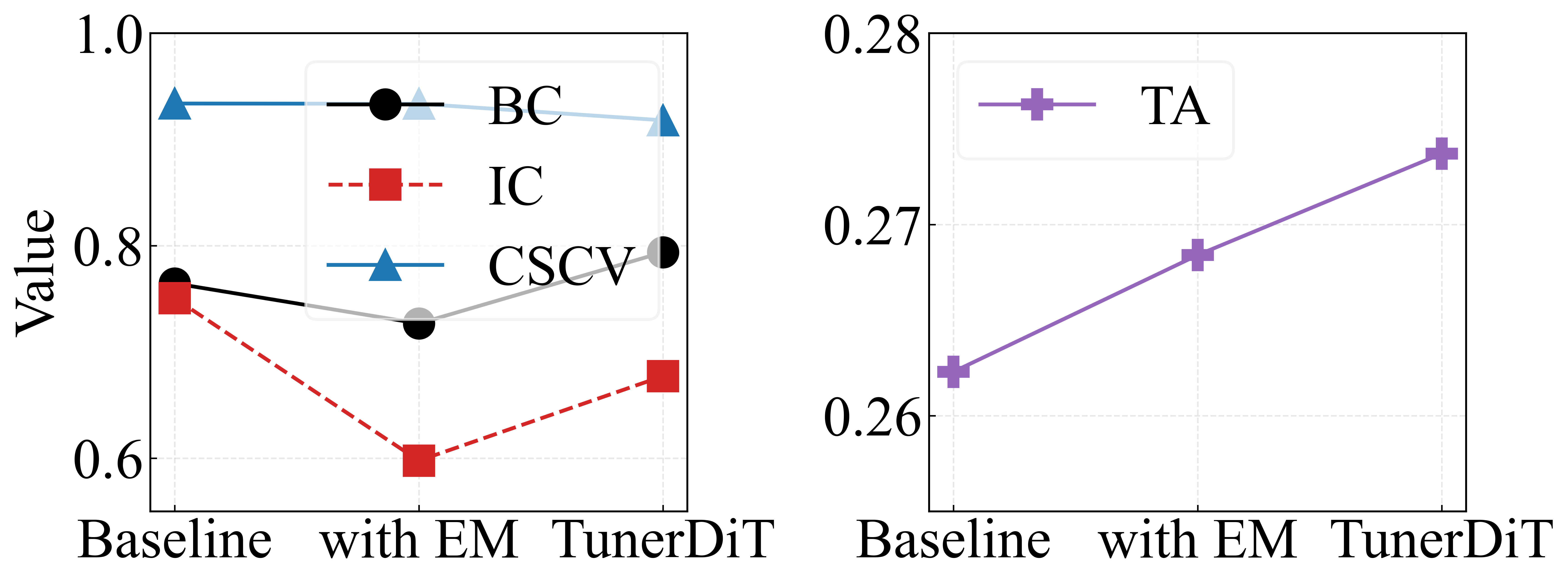}
    \caption{Metric trends of \method{} under different module combinations.}
    \label{fig:ablation_plot}
\end{figure}





Figure~\ref{fig:ablation_components}, ~\ref{fig:ablation_plot} illustrate how each steering component in \method{} contributes to multi-event generation quality.
Overall, we observe a clear tension between \emph{text alignment} (TA, TIS) and \emph{visual consistency} (BC, IC, CSCV): improving one group often degrades the other. But \method manages to achieve a \textit{balance}.

\paragraph{Effect of Event-Partitioned Mask (EM).}
As shown in Figure~\ref{fig:ablation_components}, enabling EM substantially improves multi-event compliance.
However, this comes at the cost of lower BC/IC/CSCV, since boundary enforcement caused by diagonal and transition masks suppresses cross-event leakage, which inevitably harms global smoothness and consistency. The increase of TA/TIS indicates that events become more cleanly separated and better aligned with their prompts.

\paragraph{Recovering consistency with Prompt Fusion (PF).}
Adding PF in TunerDiT restores a significant portion of the lost consistency while preserving most of the gains in multi-event alignment.
In Figure~\ref{fig:ablation_plot}, this manifests as improved BC/IC/CSCV compared to EM-only configurations, with even slight improvement in TA.
PF therefore acts as a counter-balance to EM, yielding a more favorable trade-off between the two mutually inhibiting metric groups: event separation versus semantically coherent across events. More detailed discussion is positioned in Appendix~\ref{sec:app:metrics_discussion}.




\subsection{Qualitative Analysis}
Figure~\ref{fig:quali} compares multi-event sequences for a three-step cooking prompt. The first is the direct concatenation of individual generation (Concat), which directly treats events as separate prompts as individual input to the generation process and then concatenates the video outputs together, which has the worst semantic consistency. The following three are naive multi-event prompting, showcasing the inability to tackle the three challenges, and generating the wrong event count and ordering. DitCtrl has worse text alignment and consistency when the event focuses on shifting motions. Overall, our method TunerDiT achieves the best visual quality in event ordering, transition smoothness, and consistency. For more examples, please refer to Appendix \ref{sec:app:quali}.

\subsection{Discussions}
\begin{figure}[htbp!]
    \centering
    \includegraphics[width=\linewidth]{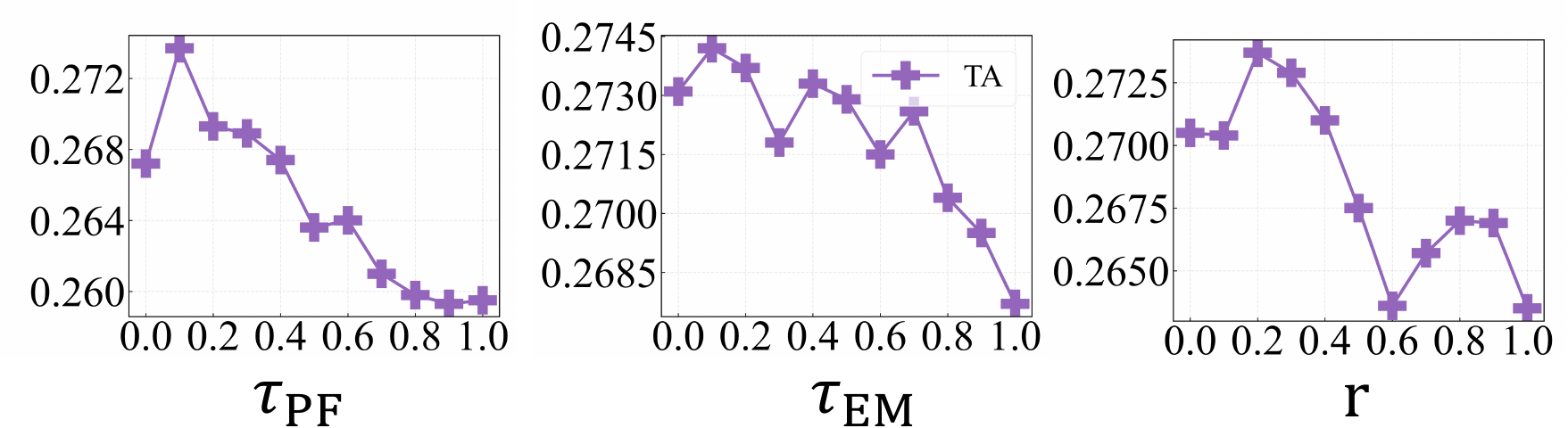}
    \caption{Effect of steering position on multi-event generation. The steering-position behavior occurs mostly during the early stage of the diffusion process.}
    \label{fig:steering_position}
\end{figure}

\begin{figure}[htbp!]
    \centering
    \includegraphics[width=\linewidth]{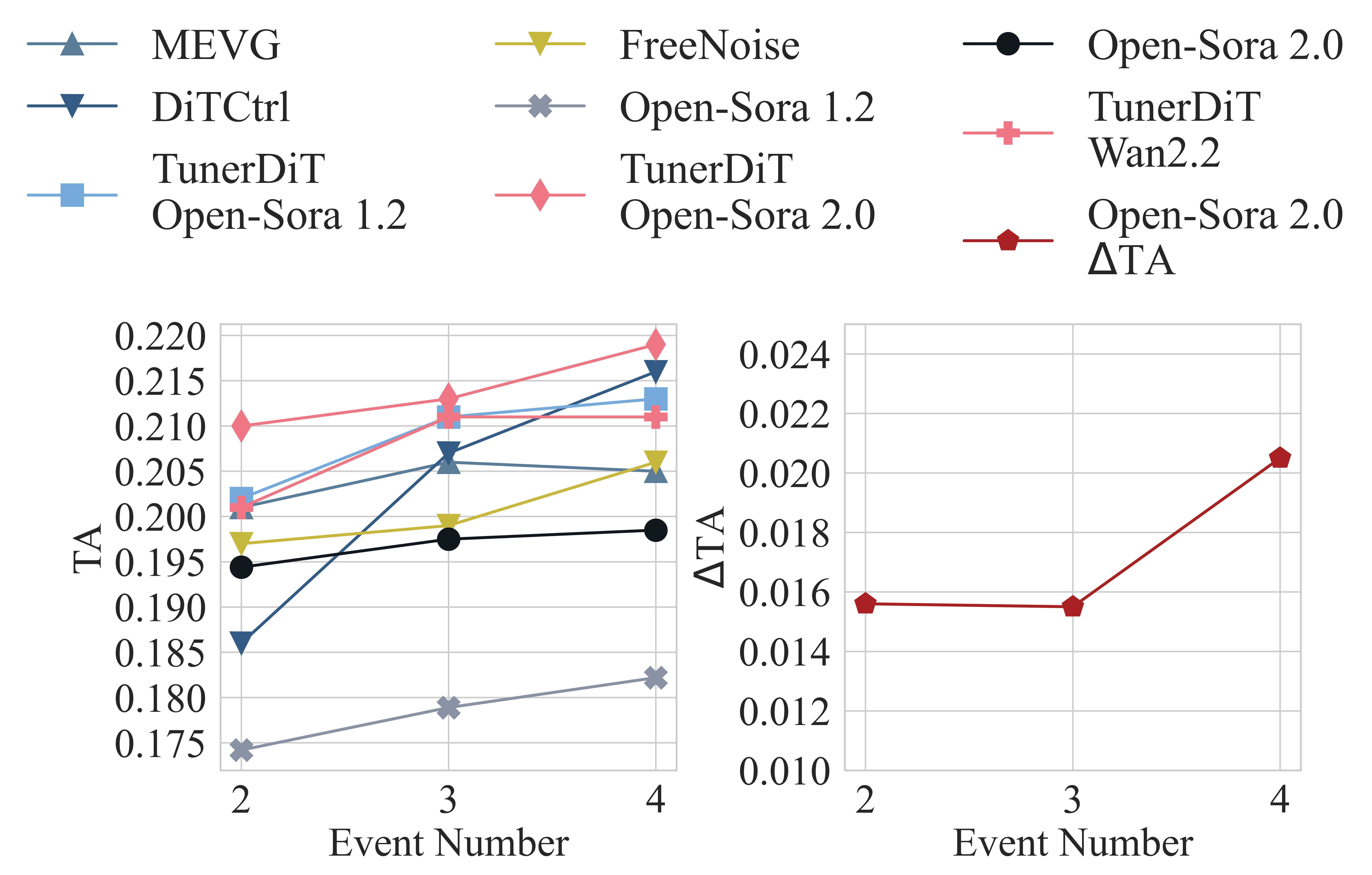}
    \caption{Performance trends of TunerDiT as the number of events increases.}
    \label{fig:eventcount}
\end{figure}


\paragraph{\textbf{Investigation on steering position and transition bandwidth ratio}.}
As shown in Fig.~\ref{fig:steering_position}, \textsc{TA} is most sensitive to the position of the prompt–fusion gate $\tau_{\text{PF}}$: performance peaks when $\tau_{\text{PF}}$ lies in the early denoising phase ($\approx\!0.0$–$0.2$ of the steps), confirming that steering the shared layout early is crucial for multi–event alignment. 
The EM steering point $\tau_{\text{EM}}$ exhibits a relatively flat plateau over a broad mid–range, however, it still peaks around $0.1-0.2$ before the very late refinement steps.
Varying the transition–band ratio $r$ further reveals that small to moderate bands ($r\approx0.1$–$0.3$) yield the best \textsc{TA} and wide bands ($r\!>\!0.6$) over–smooth boundaries that revert to the wrong behaviors of not differentiating events.
We therefore adopt an early $\tau_{\text{PF}}$, a mid–range $\tau_{\text{EM}}$, and a moderate $r$ as default, which jointly give the strongest multi–event alignment.

\paragraph{\textbf{Increased performance on more event counts}}

Figure~\ref{fig:eventcount} shows that $\mathrm{TA}(E)$ increases as $E$ grows from two to four.
Besides, we define the improvement gain over the corresponding base video diffusion models as
$\Delta\mathrm{TA}(E)=\mathrm{TA}_{\text{TunerDiT}}(E)-\mathrm{TA}_{\text{base}}(E).$
Figure~\ref{fig:eventcount} indicates that $\Delta\mathrm{TA}(E)$ increases with $E$. 
This suggests that the proposed steering is increasingly necessary at longer horizons.

We attribute this trend to the factor that more events introduce more explicit temporal anchors and boundaries, which our steering modules exploit to stabilize ordering and reduce leakage across segments; 

\paragraph{Additional Analysis} Additional thorough discussions, such as failure cases, investigation on event types, limitations, and future works, are in Appendix \ref{sec:app:additionalapp}. In-depth analysis of probing the intrinsic turning points is in ~\cite{liao2025eventsswitchmultieventvideo}.

\section{Conclusion}


We presented TunerDiT, a training-free steering progressive framework for multi-event T2V generation by utilizing the intrinsic turning point where text guidance in DiTs shifts from global layout to fine details for better text-video alignment, smooth transitions, and content consistency.
{
    \small
    \bibliographystyle{ieeenat_fullname}
    \bibliography{main}
}

\clearpage
\setcounter{page}{1}
\maketitlesupplementary

\appendix
\section{\textsc{Meve}: Benchmark Construction Details.}
\label{sec:app:dataset}

\paragraph{\textbf{Existing Multi Prompt Benchmarks}}

We compare a large body of recent work focusing on building \emph{general-purpose} video generation benchmarks as shown in Figure \ref{tab:benchmark_comparison}.
VBench~\citep{huang2023vbenchcomprehensivebenchmarksuite} decomposes “video generation quality’’ into 16 hierarchical dimensions.
VBench++~\citep{huang2024vbenchcomprehensiveversatilebenchmark} further extends this line by supporting image-to-video settings by introducing an image suite. VBench-2.0~\citep{zheng2025vbench20advancingvideogeneration} shifts the focus from superficial visual plausibility to intrinsic faithfulness.
These benchmarks are invaluable for diagnosing \emph{overall} model quality in a single semantic unit.



For \emph{multi-prompt} video generation, ShotAdapter~\citep{kara2025shotadapter} uses a general ChatGPT curated dataset, which lacks discussion for different prompt types. DiTCtrl~\citep{cai2025ditctrlexploringattentioncontrol} introduces MPVBench, a dedicated benchmark for evaluating multi-prompt transitions with ten transition types ~\citep{cai2025ditctrlexploringattentioncontrol} and represents an important first step toward benchmarking multi-prompt video generation. However, MPVBench primarily focuses on evaluating the transition effects between two similar scenes that don't differ significantly in motion content. It deliberately avoids logical sequential prompts; hence, this benchmark is not generalizable to real-world scenarios where temporal and logical events occur. We follow the definition of a single event unit that consists of four basic components, \textit{i.e., subject, predicate, objects, and time~\citep{10.1162/neco_a_01552}}, and prior benchmarks do not explicitly address this definition and have a mixture of multi-scene with multi-event settings, such as WorldScore~\citep{duan2025worldscore}, which misses the basic four components for an event.


\paragraph{\textbf{Motivation and \textsc{Meve} Construction Details}}

In contrast, our goal with \textsc{MEve} is to provide a targeted benchmark for \emph{multi-event} video generation, built from three complementary sources:
(1) conventional LLM-generated prompts tailored to multi-event stories;
(2) multi-event variants of existing general video benchmarks (e.g., VBench 2.0~\citep{zheng2025vbench}) obtained by transforming single-scene prompts into multi-event counterparts; and
(3) real-world event sequences derived from egocentric videos with natural-language narrations, which supply temporally and logically coherent event chains, further supported by paired ego–exo views. Besides, all categories are carefully organized into 2-, 3-, and 4-event prompt settings with explicit event segmentation and ordering, with extensibility to more. This design makes \textsc{MEve} novel and complementary to previous benchmarks. Specifically, we construct \textsc{MEve} as follows:


\paragraph{LLM as data synthesizer}
Following prior settings in \cite{kara2025shotadapter}, we use Gemini 2.5 Pro ~\cite{comanici2025gemini25pushingfrontier} as a controlled prompt generator and author $60$ event narratives (e.g., ``$\texttt{event}_1 \; \text{then} \; \texttt{event}_2$ \; \text{then} \; \dots''). The prompts target the \emph{General} domain and are written to avoid specific scene priors or content biases.

\paragraph{Prompts of diagnostic content}
To inherit general evaluation while disentangle how subject and action factors shape multi-event generation, we adapt VBench~2.0 categories~\cite{zheng2025vbench, zhang2024evaluationagent} that are critical in multi-event settings. Specifically, we convert prompts from categories \emph{Motion Order Understanding}, \emph{Human Identity}, and \emph{Complex Plot} into four sequential events per prompt while preserving each category's diagnostic intent.

\paragraph{Prompts of real-world reference}
To maintain real-world temporal and logical coherence between events and isolate the effect of \emph{viewpoint} on event controllability, we construct paired egocentric and exocentric variants of prompts from video narrations from Ego–Exo4D~\citep{grauman2024egoexo4dunderstandingskilledhuman}. 
For viewpoint control, each prompt is prepended with one of two textual formats:  (a) \textit{Egocentric:} Add the prefix ``Generate a first-person view video...'', where the original subject $C$ is rephrased as ``the camera wearer''. (2) \textit{Third-person:} Add the prefix ``Generate a third-person view video...'', denoting an external viewpoint, where the original subject $C$ is transformed into ``the person''.  
This process results in $50$ pairs of prompts, yielding in total $100$ prompts.


\section{Implementation Details}
\label{sec:appendix_implementation}


Under the restriction of the same GPU resource, the number of frames of videos in different event counts and the resolution of the generated videos are shown in Table \ref{tab:model_resolution_comparison}.
\begin{table}[h!]
\centering
\caption{Number of frames of different event count and frame resolution}
\label{tab:model_resolution_comparison}
\begin{tabular}{lcccc}
\toprule
\textbf{Model Name} & \textbf{\#2} & \textbf{\#3} & \textbf{\#4} & \textbf{Resolution} \\ \hline
Open-Sora 1.2      & 173        & 201        & 205        & 1280$\times$720     \\
Open-Sora 2.0      & 204        & 204        & 204        & 256$\times$256      \\
Wan2.2 & 97 & 97 & 97 &  1280$\times$720 \\
MEVG              & 32         & 48         & 64         & 256$\times$256      \\
DiTCtrl           & 97         & 145        & 193        & 720$\times$ 480    \\
FreeNoise & 192 & 192 & 192 & 256$\times$ 256 \\
\bottomrule
\end{tabular}
\end{table}

\paragraph{Event Partitioned Mask}

OpenSora 1.2 employs ST-DiT with separate temporal and spatial blocks for self-attention and a shared cross-attention layer between text embedding and video embedding. OpenSora 2.0 utilizes a dual-stream DiT architecture that concatenates text and video embeddings before a unified self-attention. This self-attention on the new embedding incorporates cross-attention, video self-attention, and text self-attention simultaneously. 
For ST-DiT in OpenSora 1.2, the Event Partitioned Mask is applied exclusively to the temporal blocks. For dual-stream DiT in OpenSora 2.0, the Event Partitioned Diagonal Mask operates on both blocks. 

\textit{Turning points and onset steps.} The Event Partitioned Mask is not adopted in previous $\tau$ denoising steps, and is adopted only in the remaining denoising steps. 
For TunerDiT Open-Sora 1.2, the turning point $x=0.1$ and $\tau_{\text{EM}}=\lfloor 0.1\times N \rfloor $. For TunerDiT Open-Sora 2.0 and Wan 2.2, $x=0.2$ and $\tau_{\text{{EM}}}=\lfloor 0.2\times N \rfloor $. 

\textit{Text embedding partitioning.} Event prompts are embedded separately and concatenated together to enable proper event segmentation and ordering through the Event Partitioned Mask. To maintain consistent positional embeddings, text embeddings are zero-padded to the maximum sequence length. 


\paragraph{Inter and inner event bandwidths.} Consider a prompt composed of $E$ distinct events. Given a video with $ F_l$ latent frames in total, a video segment of $F_l/E$ latent frames is allocated to each event. Let $r$ represent the band size ratio for band size ratio,  the corresponding transition bandwidth is calculated as $H_{\text{inter}} = W_{\text{inter}} = \frac{r F_l}{E}$. This mask is applied in TunerDiT Open-Sora 1.2 with $\beta=0.1$ and TunerDiT Open-Sora 2.0 and Wan 2.2 with $\beta=0.2$. 

\paragraph{Cross Event Prompt Fusion}


For TunerDiT Open-Sora 1.2 and TunerDiT Open-Sora 2.0, the $\lfloor \tau_{\text{on}}^{\text{fuse}} = 0.1\times N \rfloor$, where $N$ is the total number of denoising steps.




\section{Human Evaluation}
\label{sec:app:humaneval}

\paragraph{Evaluation Details}
Following human evaluation settings in \cite{cai2025ditctrl}, to ensure a comprehensive evaluation, we selected samples with three distinct event counts of 2, 3, and 4 from six different datasets, for a total of 18 prompts. For each prompt, we choose one sample across 8 models, yielding a total of $8 \times 3 \times 6 = 144$ videos for human assessment.
As shown in our interface Figure \ref{fig:interface}. All models are anonymized and randomly shuffled. Participants are asked to rate the following four metrics on a 5-point Likert scale, where 1 indicates the lowest satisfaction and 5 indicates the highest, for each video  :

\begin{figure}[htbp!]
    \centering
    \includegraphics[width=1\linewidth]{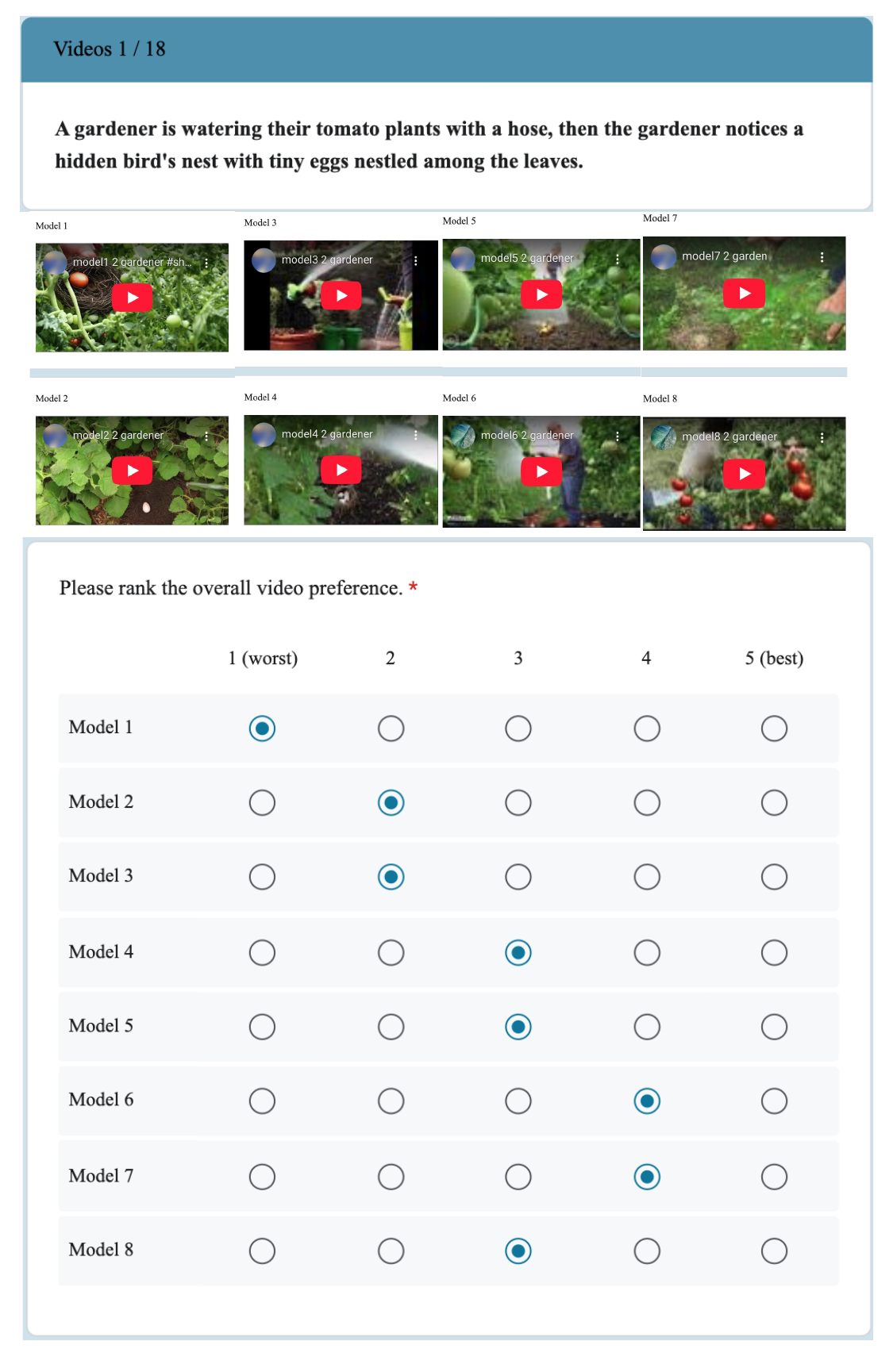}
    \caption{Interface of annotating multi-event generation quality.}
    \label{fig:interface}
\end{figure}

\begin{enumerate}
    \item Overall Preference: \textit{Please rate your overall preference for the video.}
    \item Motion Naturalness: \textit{How natural and realistic is the motion in the video?}
    \item Transition Smoothness: \textit{How smoothly does the video content transition across different frames?}
    \item Text-Video Alignment: \textit{To what extent does the video content match the given text descriptions?}
\end{enumerate}

\paragraph{Evaluation Results}
The quantitative results, derived from evaluations by 14 participants with mimum degree in bachlors, are presented in Table \ref{tab:merged_vla_human}. 
TunerDiT achieves superior Text-Video Alignment scores compared to the baseline models, including open-source base models and zero-shot methods MEVG, DiTCtrl and FreeNoise. 
Furthermore, TunerDiT exhibits strong improvement over its base models with high performance gains across all questions, validating the overall acceptance by human evaluators.

\section{Discussion on Metrics}
\label{sec:app:metrics_discussion}



\paragraph{TA and TIS: \textbf{Frame-wise and Shot-wise}}
But base models that by default generate fuse events blending backgrounds and subjects from multiple events across the entire video can obtain spuriously high Text alignment (TA) and text–image similarity (TIS) scores, due to each frame representation being conditioned on all events rather than cleanly separated event-specific content. Therefore, it is necessary to emphasize both metrics: TA, which is computed with ViCLIP on a shot-wise alignment, and TIS, which is computed with frame-wise alignment. This explains the fact that our model has greater improvement over TA.

\paragraph{BC, IC, and CSCV: \textbf{Consistency Metric Hacking}}

These default settings that models could not distinguish events in generation influence the BC, IC, and CSCV scores severely. Fusing every cue in the prompts on every frame, regardless of the temporal event order, will lead to the highest consistency score, while breaking generation into event segments inevitably lowers it, although more semantically correct multi-event videos are generated and aligned to human evaluation.  We define this phenomenon as \emph{Consistency Hacking}, that blending, or generally static backgrounds and subjects, increases CSCV due to minimal inter-frame change as well as BC and IC due to unchanged appearance, even though the video fails at multi-event separation. 

Metric hacking is further exacerbated by methods that can separate events without deliberately noticing this. For example, Mask2DiT\citep{qi2025mask2ditdualmaskbaseddiffusion}, although generating events separately, generates backgrounds and subjects that remain largely static in each event. This further indicates that BC, IC, and CSCV scores do not necessarily decrease for ideal multi-event videos.

\paragraph{\textbf{Tradeoff between Alignment and Consistency }}. Given the ablation study in Figure \ref{fig:ablation_components}, \ref{fig:ablation_plot}, the incorporation of Prompt Fusion facilitates smoother motion dynamics during event transitions within the video. However, it does not necessarily improve TA scores, since frames sampled from transition segments between consecutive events align semantically less to both the preceding and the subsequent event, thereby it trades alignment with consistency. 


\begin{figure}[t]
    \centering
    \includegraphics[width=1\linewidth]{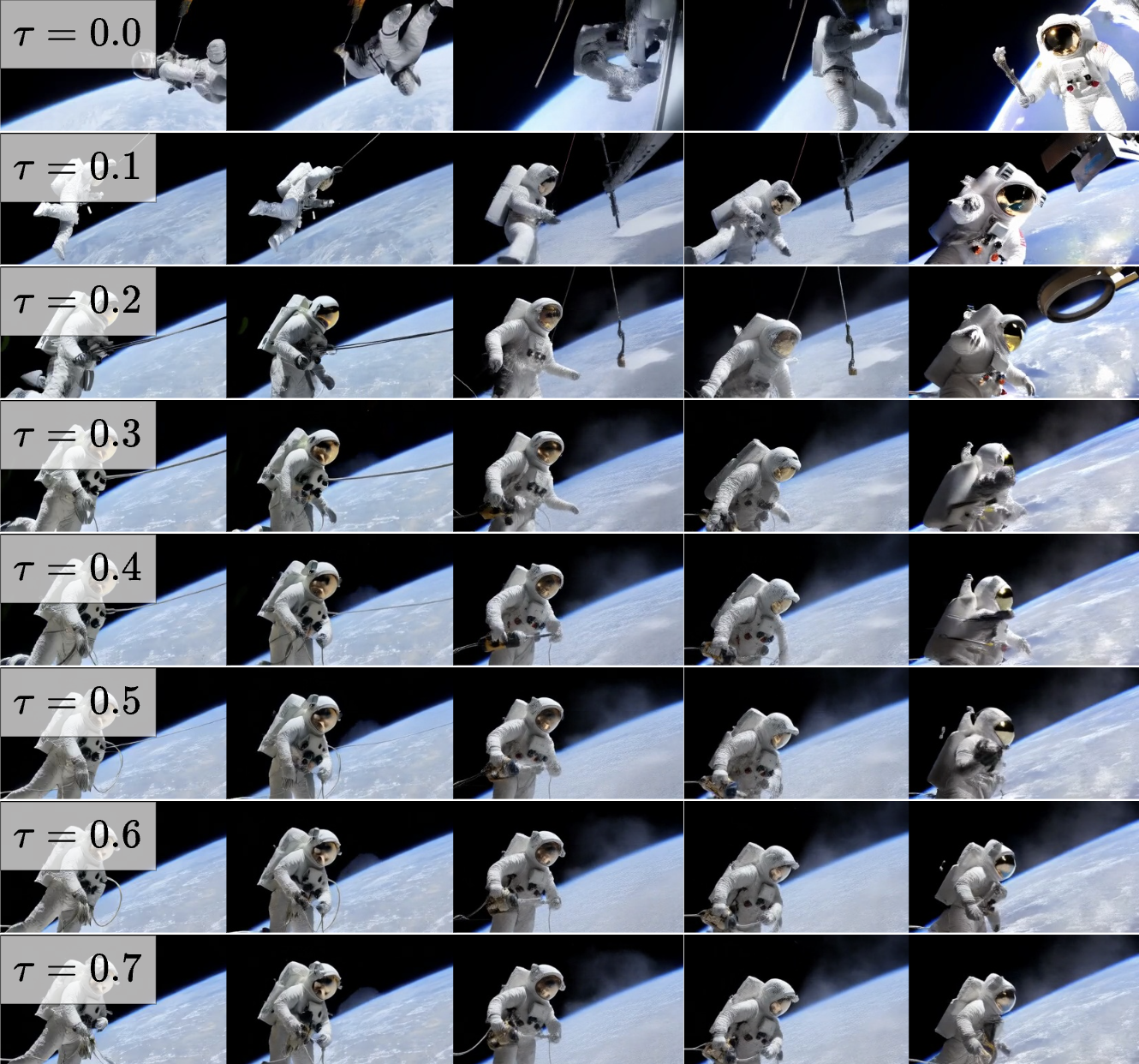}
    \caption{Visualization of generated frames as the Event Partitioned Mask ratio increases from $0.0$ to $0.7$.}
    \label{fig:ratio_visualization}
\end{figure}

\section{More Qualitative Analysis}
\label{sec:app:quali}
As shown in Figure \ref{fig:ratio_visualization}, we visualize the event generation conditioned on distinct turning point values $x$ of the Event Partitioned Mask (EM) over a fixed grid
$\mathcal{X}=\{0,\,0.1,\,0.2,\,\ldots,\,0.7\}$. \\ Events begin to merge when $x > 0.3$ in the generated videos, reflecting the plateau effect in Figure \ref{fig:tunerbutton}.  This observation empirically confirms the finding that influence transitions of text conditions occur at approximately $30\%$ of the denoising steps in DiT models. When the Event Partitioned Mask doesn't provide enough event segmentation information because of too high $x$, multiple events merge indistinguishably, and gradually deteriorate the performance back to the original base model.

\section{Additional Discussions}
\label{sec:app:additionalapp}

\begin{figure}[t]
    \centering
    \includegraphics[width=1.0\linewidth]{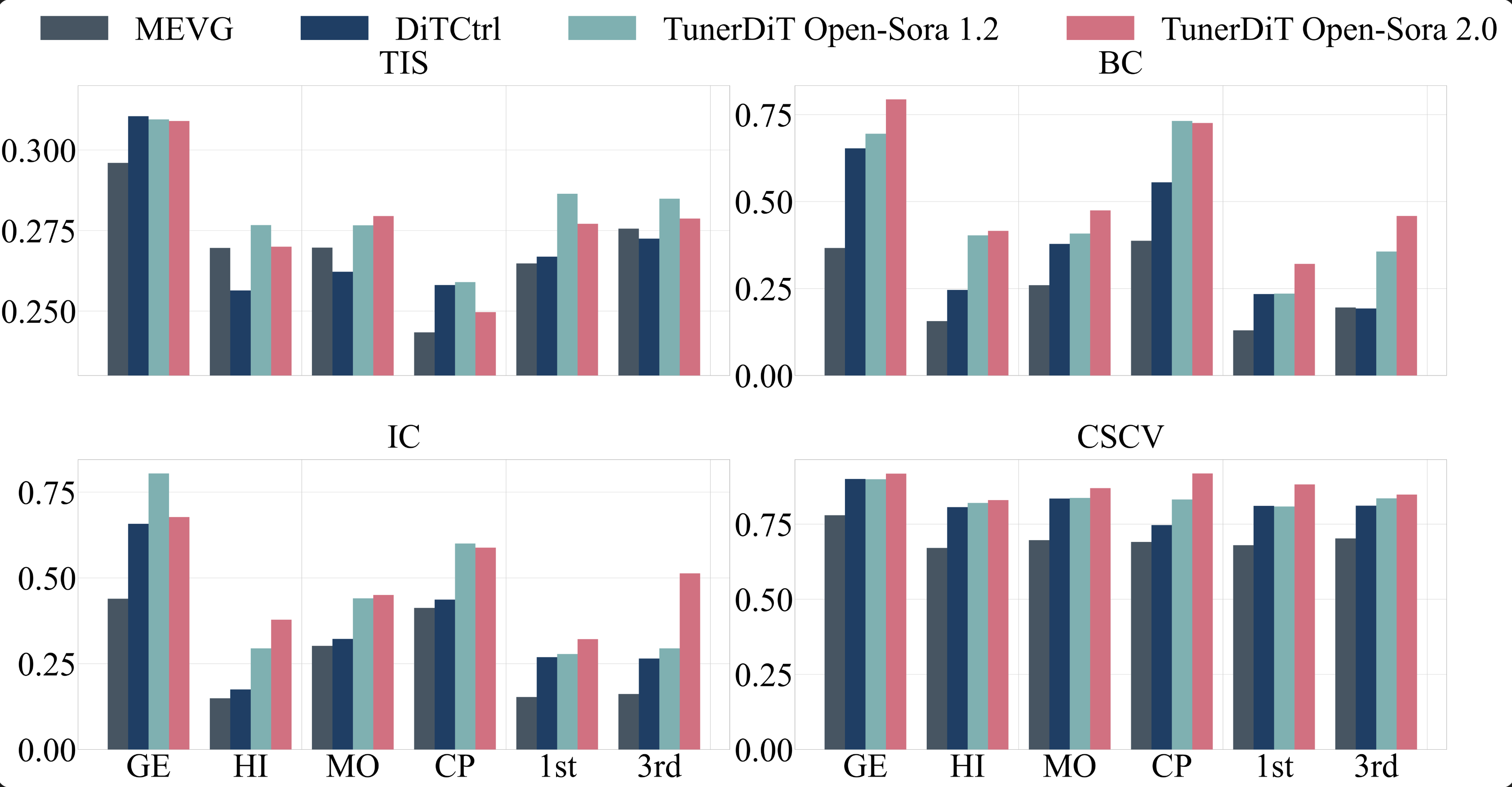}
    \caption{Investigation on dataset categories. Notion: GE (general prompts), HI (human identity), MO (motion order), CP (complex plots), and view splits (1st, 3rd). }
    \label{fig:datasets}
\end{figure}

\paragraph{\textbf{Investigation on Dataset Categories}}

We compare zero-shot methods with our \emph{TunerDiT} on Open\mbox{-}Sora\,1.2 and Open\mbox{-}Sora\,2.0 as shown in Fig.~\ref{fig:datasets}.
TunerDiT ranks first on nearly all bars in BC and CSCV and shows consistent gains in TIS. IC and BC see the largest gains. Event-partitioned masking preserves per-event identity while prompt fusion stabilizes late-stage detail, yielding higher person and background consistency.  
\emph{For motion order (MO)}, TIS increases, but BC/IC gains are moderate, which is consistent with the fact that MO stresses temporal ordering more than appearance persistence. 
\emph{From viewpoint control} Generation from both views benefits from TunerDiT. 1st-person videos show marked IC improvements due to consistent hands and objects, while 3rd-view shows strong BC and CSCV due to more stable backgrounds.

\paragraph{\textbf{Implications and guidance.}}
(1) The largest gains occur in HI and CP, where boundary control and scene-conditioned bands are most critical;  
(2) For MO-heavy prompts, later-onset prompt fusion is effective, while mask ratio can be reduced to avoid over-isolation;  
(3) The consistent CSCV advantage indicates that local cross-event bands are sufficient for smooth transitions even under camera changes. 

\paragraph{\textbf{Failure Cases}}
We identify two primary failure modes in TunerDiT: \textit{excessive fusion} and \textit{insufficient fusion}. The former yields results resembling the original baseline models, where multiple events become indistinguishable or erroneously blended into a single sequence. Conversely, the latter leads to semantic discontinuities, such as inconsistent subjects or backgrounds across event boundaries, producing abrupt changes similar to those observed in the Concat baseline.

\begin{figure}[htbp!]
    \centering
    \includegraphics[width=1\linewidth]{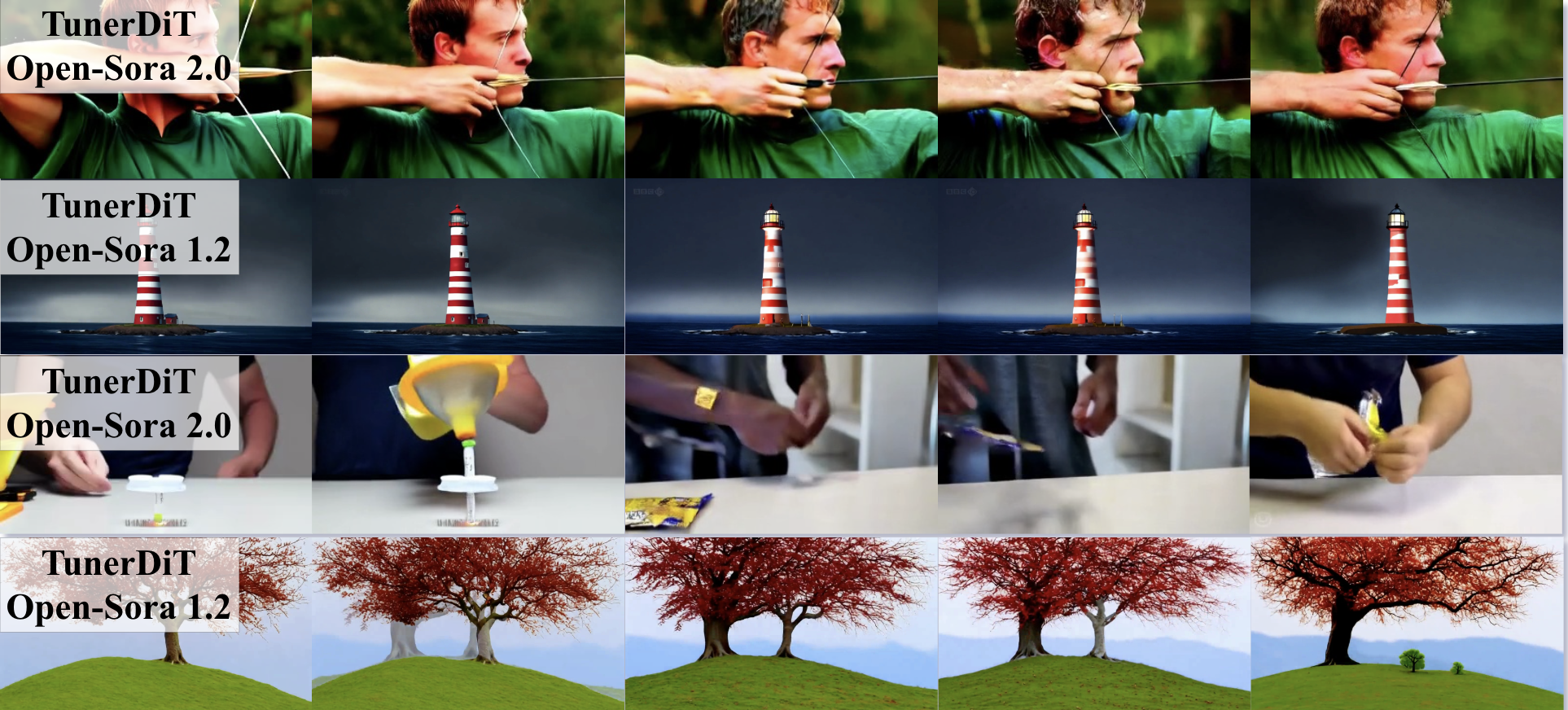}
    \caption{Visualization of failure modes in TunerDiT caused by extreme event fusion settings. The upper two rows demonstrate the effect of \textit{\textbf{excessive fusion}}, where distinct events collapse into a single, static sequence. Conversely, the lower rows illustrate \textbf{\textit{insufficient fusion}}, resulting in temporal inconsistencies regarding the subject and background across event boundaries.}
    \label{fig:failure_cases}
\end{figure}

\paragraph{\textbf{Limitations and Future Works}}
The current implementation is constrained by the maximum text embedding length of the baseline model architecture. Given a fixed text embedding window of 512 tokens currently adopted by Open-Sora 2.0, a prompt describing $E$ events implies that each event is allocated an average of only $\frac{512}{E}$ tokens. This token number limits the method's scalability to extremely complex sequences. However, theoretically, assuming better computational situations and the rapid development of video diffusion models for long generation, this framework can be applied naturally to longer videos with more distinct events in future due to its steering method that originates from the intrinsic perspectives shared by DiT models. 

\section{VLM Judgement}

\paragraph{VLM-as-a-judge evaluation.}
Tables~\ref{tab:vla_ei_breakdown} and ~\ref {tab:vla_tva_breakdown} report the category-wise breakdown of Event Isolation (EI) and Text--Video Alignment (TVA) evaluated by a VLM judge. Overall, \method{} Open-Sora~2.0 achieves the strongest average performance on both metrics, with the \textit{highest} EI score of \textbf{$0.572$} and the \textit{highest} TVA score of \textbf{$1.533$}, indicating that it best separates multiple events while maintaining the strongest semantic alignment to the input prompt. 

On EI, \method{} Open-Sora~2.0 performs particularly well in the \textit{Complex Plot} and ego/exocentric view categories (\textit{First}, \textit{Third}), where it reaches $0.8$, $0.7$, and $0.7$, respectively, substantially outperforming all zero-shot baselines. \method{} Open-Sora~1.2 also shows strong EI on the \textit{HI}, \textit{First}, and \textit{Third} categories, suggesting that the proposed steering is especially beneficial when subject identity and viewpoint changes must still preserve event separability. 

On TVA, the three \method{} variants consistently outperform the zero-shot baselines in average score, confirming that better event isolation does not come at the cost of global prompt alignment. In particular, \method{} Open-Sora~2.0 obtains the best or tied-best results in \textit{General}, \textit{HI}, and \textit{MO}, while \method{} Wan2.2 is competitive in the viewpoint-sensitive \textit{Third} category. 

These results show that \method{} improves both the \emph{structural correctness} of the multi-event setting, as measured by EI, and the \emph{semantic faithfulness} of the resulting videos, as measured by TVA, across diverse prompt categories.

\begin{table*}[t]
\centering
\caption{VLA as judge score on Event Isolation (EI). Evaluating each frame, infer the number of events, then compare with the actual number of events in the prompt.}
\begin{tabular}{l|c|cccccc}
\hline
\textbf{Name} & \textbf{Average} & \textbf{General} & \textbf{HI} & \textbf{MO} & \textbf{CP} & \textbf{First} & \textbf{Third} \\
\hline
\multicolumn{8}{l}{\textbf{Zero-shot Methods}} \\
\hline
MEVG        & 0.435 & 0.315 & 0.6  & 0.5  & 0.7  & 0.3   & 0.2  \\
FreeNoise   & 0.436 & 0.67  & 0.44 & 0.44 & 0.44 & 0.375 & 0.25 \\
DitCtrl     & 0.375 & 0.15  & 0.5  & 0.2  & 0.3  & 0.5   & 0.6  \\
Maks2DiT    & 0.108 & 0.05  & 0.1  & 0.0  & 0.2  & 0.2   & 0.1  \\
VideoCrafter2 & 0.08 & 0.1  & 0.0  & 0.0  & 0.1  & 0.2   & 0.1  \\
\hline
\multicolumn{8}{l}{\textbf{Ours}} \\
\hline
Tuner Wan2.2 & 0.474 & 0.4  & 0.43 & 0.32 & 0.5 & 0.6 & 0.5 \\
Tuner Open-Sora 1.2 & 0.499 & 0.2 & 0.6  & 0.3  & 0.5 & 0.8 & 0.6 \\
Tuner Open-Sora 2.0 & \textbf{0.572 }& 0.33 & 0.5  & 0.4  & 0.8 & 0.7 & 0.7 \\
\hline
\end{tabular}
\label{tab:vla_ei_breakdown}
\end{table*}

\begin{table*}[t]
\centering
\caption{VLA as judge score on Text-Video Alignment (TVA). Evaluating each frame, generating captions, and then comparing the alignment score between the captions and the original video prompt.}
\begin{tabular}{l|c|cccccc}
\hline
\textbf{Name} & \textbf{Average} & \textbf{General} & \textbf{HI} & \textbf{MO} & \textbf{CP} & \textbf{First} & \textbf{Third} \\
\hline
\multicolumn{8}{l}{\textbf{Zero-shot Methods}} \\
\hline
Mask2DiT     & 1.275 & 1.95 & 1.2 & 1.0 & 1.3 & 1.0 & 1.2 \\
DitCtrl      & 1.425 & 2.25 & 1.5 & 1.5 & 1.2 & 1.1 & 1.0 \\
MEVG         & 1.375 & 1.35 & 1.8 & 1.4 & 1.3 & 1.0 & 1.4 \\
Freenoise    & 1.4   & 1.8  & 1.7 & 1.5 & 1.0 & 1.0 & 1.4 \\
VideoCrafter2 & 1.183 & 1.5  & 1.1 & 1.2 & 1.3 & 1.0 & 1.0 \\
\hline
\multicolumn{8}{l}{\textbf{Ours}} \\
\hline
Tuner Wan2.2 & 1.503  & 1.72 & 1.6 & 1.8 & 1.2 & 1.2 & 1.5 \\
Tuner Open-Sora 1.2   & 1.492 & 1.95 & 1.5 & 1.6 & 1.2 & 1.3 & 1.4 \\
Tuner Open-Sora 2.0 & \textbf{1.533}  & 2.1  & 1.8 & 1.8 & 1.0 & 1.4 & 1.1 \\
\hline
\end{tabular}
\label{tab:vla_tva_breakdown}
\end{table*}

\end{document}